\documentclass{article}

\usepackage{microtype}
\usepackage{graphicx}
\usepackage{multirow}
\usepackage{multicol}
\usepackage{float}
\usepackage{graphicx}
\usepackage{subfigure}
\usepackage{booktabs} 
\usepackage[table,xcdraw]{xcolor}
\usepackage{tcolorbox}
\usepackage{fancyvrb}
\usepackage{fvextra}
\definecolor{cGreen}{HTML}{2e75b5}
\definecolor{cgray}{HTML}{FAFAFA}
\definecolor{blue}{HTML}{0055cc}
\definecolor{red}{HTML}{cc1100}
\definecolor{orange}{HTML}{cc7700}
\definecolor{green}{HTML}{339955}
\definecolor{Highlight}{rgb}{0.12,0.49,0.85}
\definecolor{my_red}{HTML}{ff0000}
\definecolor{brickred}{rgb}{0.8, 0.25, 0.33}
\definecolor{brandeisblue}{rgb}{0.0, 0.44, 1.0}
\definecolor{blueish}{rgb}{0.0, 0.3, 0.6}
\definecolor{darkgreen}{rgb}{0, 0.5, 0}
\definecolor{tab_green}{RGB}{226, 239, 217}
\definecolor{tab_blue}{RGB}{251, 229, 213}

\usepackage{hyperref}

\usepackage[preprint]{icml2026}

\usepackage{amsmath}
\usepackage{amssymb}
\usepackage{mathtools}
\usepackage{amsthm}
\usepackage{url}
\usepackage{caption}
\usepackage{wrapfig}
\usepackage[capitalize,noabbrev]{cleveref}
\usepackage{pifont}
\setcitestyle{circle}
\crefname{section}{Sec.}{Secs.}
\Crefname{section}{Section}{Sections}
\Crefname{table}{Table}{Tables}
\crefname{table}{Tab.}{Tabs.}
\crefname{equation}{Eq.}{Eqs.}
\Crefname{equation}{Eq.}{Eqs.}
\crefname{figure}{Fig.}{Figs.}
\Crefname{figure}{Fig.}{Figs.}
\crefname{algorithm}{Algo.}{Algo.}
\Crefname{algorithm}{Algo.}{Algos.}
\usepackage{xspace}

\newcommand{\NAME}{Dummy Forcing\xspace}
\newcommand{\suppl}{\textit{Appendix}\xspace}
\usepackage[capitalize,noabbrev]{cleveref}

\theoremstyle{plain}

\theoremstyle{definition}

\theoremstyle{remark}

\usepackage{soul}
\usepackage{arydshln}
\usepackage[textsize=tiny]{todonotes}

\icmltitlerunning{Efficient Autoregressive Video Diffusion with Dummy Head}

\begin{document}
\twocolumn[
   \icmltitle{Efficient Autoregressive Video Diffusion with Dummy Head}
  \icmlsetsymbol{equal}{*}
  \icmlsetsymbol{intern}{$\dagger$}
  \begin{icmlauthorlist}
    \icmlauthor{Hang Guo}{thu,msra}
    \icmlauthor{Zhaoyang Jia}{msra,ustc}
    \icmlauthor{Jiahao Li}{msra}
    \icmlauthor{Bin Li}{msra}
    \icmlauthor{Yuanhao Cai}{jhu}
    \icmlauthor{Jiangshan Wang}{thu}
    \icmlauthor{Yawei Li}{ethz}
    \icmlauthor{Yan Lu}{msra}
  \end{icmlauthorlist}

  \icmlaffiliation{thu}{Tsinghua University}
  \icmlaffiliation{ethz}{ETH Z\"{u}rich}
  \icmlaffiliation{msra}{Microsoft Research Asia}
  \icmlaffiliation{jhu}{Johns Hopkins University}
  \icmlaffiliation{ustc}{University of Science and Technology of China}
  \icmlcorrespondingauthor{}{}
  \icmlkeywords{Video Generation, Autoregressive Diffusion Models, KV Cache Pruning}
  \vskip 0.3in
]
\printAffiliationsAndNotice{}  
\begin{abstract}
The autoregressive video diffusion model has recently gained considerable research interest due to its causal modeling and iterative denoising. In this work, we identify that the multi-head self-attention in these models under-utilizes historical frames: approximately 25\% heads attend almost exclusively to the current frame, and discarding their KV caches incurs only minor performance degradation. Building upon this, we propose Dummy Forcing, a simple yet effective method to control context accessibility across different heads. Specifically, the proposed heterogeneous memory allocation reduces head-wise context redundancy, accompanied by dynamic head programming to adaptively classify head types. Moreover, we develop a context packing technique to achieve more aggressive cache compression. Without additional training, our Dummy Forcing delivers up to \textbf{2.0$\times$} speedup over the baseline, supporting video generation at \textbf{24.3 FPS} with less than 0.5\% quality drop. Project page is available at \url{https://csguoh.github.io/project/DummyForcing/}.
\end{abstract}

\section{Introduction}

Video diffusion models~\cite{brooks2024sora,peebles2023dit,wan2025wan} have achieved remarkable progress in recent years, with the state-of-the-art models now being able to generate realistic shots with complex motions. Early works typically rely on bidirectional attention in diffusion transformers to generate all video frames at once. Under this paradigm, users have to wait until the model processes all frames before they can watch the video, which is slow and prevents interactive generation.

Recently, autoregressive video diffusion models~\cite{chen2024diffusionforcing,yin2025causvid,huang2025selfforcing} have shifted the paradigm to frame-by-frame generation, where each autoregressive step iteratively denoises to produce clean frames. Unlike previous bidirectional counterparts, autoregressive video diffusion supports a caching mechanism and can aggregate historical context stored in the KV cache through self-attention. By combining the strengths of autoregressive modeling and diffusion denoising, current methods allow sequential frame modeling while maintaining high visual quality. However, existing methods still face efficiency challenges when processing long visual token sequences. For instance, the KV cache length for past frames increases significantly in computation-dense tasks such as long videos or high-resolution videos. Although most current works~\cite{yang2025longlive,liu2025rollingforcing} improve efficiency at the input level by employing a sliding window strategy to restrict the model's attention only to recent frames, the model's internal utilization on contextual frames still remains a black box and has been largely unexplored.

\begin{figure}[!tb]
    \centering
    \includegraphics[width=\linewidth]{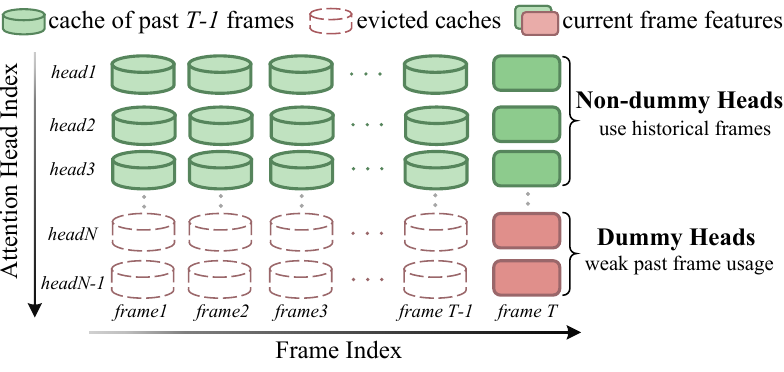}
    \vspace{-7mm}
    \caption{Weak context utilization in the multi-head attention of existing methods, e.g., Diffusion Forcing~\cite{chen2024diffusionforcing}, Self Forcing~\cite{huang2025selfforcing}. Naively pruning all KV caches of 25\% heads results in only a marginal performance drop (84.0 vs. 83.78) while speedup inference from 17.6FPS to 19.6FPS.}
    \vspace{-3mm}
    \label{fig:intro}
\end{figure}

\begin{figure*}[!tb]
    \centering
    \includegraphics[width=\linewidth]{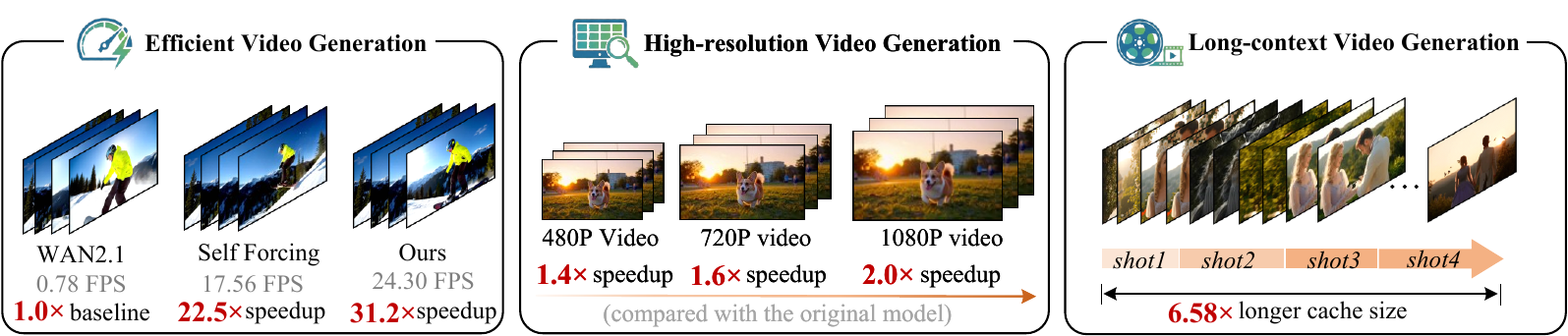}
    \vspace{-5mm}
    \caption{The proposed \NAME can be applied to (1) efficiently generate videos, (2) overcome quadratic complexity in high-resolution video generation, and (3) enlarge context lengths without increasing computational overhead.}
    \label{fig:teaser}
    \vspace{-4mm}
\end{figure*}

To this end, we delve deep into the multi-head self-attention layers of existing models, as they are responsible for context aggregation. Surprisingly, we found most autoregressive video diffusion pipelines~\cite{chen2024diffusionforcing,huang2025selfforcing,liu2025rollingforcing} are inefficient in context utilization: about 25\% attention heads disproportionately allocate large attention scores (over 80\%) to the current frame, even with access to historical frames (\cref{sec:motivation} gives a detailed discussion). Furthermore, as shown in~\cref{fig:intro}, we remove the KV cache corresponding to these heads during inference, and this simple modification causes only a 0.26\% performance drop. These observations suggest that the pre-trained models have learned a shortcut to prioritize certain attention heads to perform context aggregation, while leaving the other heads to mainly refine the current frame. Since the latter heads do not actually work on context aggregation, we refer to them as ``\textbf{dummy heads}'' in this paper.

Based on the above insight, we propose \NAME to compress redundant contextual information for efficient autoregressive video diffusion models. Specifically, we develop the Heterogeneous Memory Allocation (HMA), which assigns adaptive context lengths based on distinct head types to eliminate redundant historical frames. Subsequently, we introduce Dynamic Head Programming (DHP), which maximizes an information retention objective and uses a dynamic programming algorithm to derive optimal head classification results. Finally, we employ Packed Attention Forward (PAF) to achieve more aggressive dummy head numbers through adjust the classification boundary between dummy and non-dummy heads.

In short, we make the following key contributions:

\textbf{I.} We identify that current autoregressive video diffusion models exhibit inefficient context utilization. Furthermore, we observe dummy heads in these models whose attentions are exclusively focused on the current frame, even though past frames are available.

\textbf{II.} We introduce \NAME to efficiently compress the redundant context of dummy heads. The proposed method enables adaptive cache management and head classification while achieving high compression ratios.

\textbf{III. } As shown in~\cref{fig:teaser}, we apply \NAME to multiple video generation applications. Without any training, the proposed method achieves up to 2.0$\times$ end-to-end acceleration compared to the baseline and can generate videos at speeds exceeding 24 FPS.

\section{Related Work}

\noindent
\textbf{Video Diffusion Models.}
Early video diffusion models~\cite{polyak2024moviegen,chen2023videocrafter1,brooks2024sora,zhang2025show1,zhang2025framepack} typically feed all frames into the model at once and employ bidirectional attention during the denoising process. With large-scale training, these billion-parameter models have achieved impressive results~\cite{yang2024cogvideox,wan2025wan,kong2024hunyuanvideo}. However, bidirectional attention~\cite{vaswani2017attention} incurs significant quadratic complexity, making it challenging to support long video generation where the token sequence can be very long. Recently, autoregressive video diffusion models~\cite{chen2024diffusionforcing,henschel2025streamingt2v,teng2025magi,chen2025skyreels,yin2025causvid,gu2025far} have made substantial progress. By enabling streaming generation and KV caching, these methods significantly reduce the computational cost of video synthesis. CausVid~\cite{yin2025causvid} distills a bidirectional WAN~\cite{wan2025wan} into an autoregressive one for fast generation. Self Forcing~\cite{huang2025selfforcing} addresses the train–test mismatch by conditioning on frames from the model's outputs. RealTime~\cite{krea_realtime_14b} scales the parameters in Self Forcing further to 14B. LongLive~\cite{yang2025longlive} introduces KV re-caching to enable interactive video generation. SANA-Video~\cite{chen2025sana} proposes linear attention for efficient long video modeling. Rolling Forcing~\cite{liu2025rollingforcing} designs a joint denoising scheme with varying noise levels to reduce error accumulation.

\noindent
\textbf{KV Cache Compression in LLMs.}
KV cache pruning has been extensively studied in LLMs due to the inherently autoregressive nature of language~\cite{wan2024d2o,oren2024tova,qin2025cake,cai2025rkv,jiang2024minference,li2024snapkv,cai2024pyramidkv}. For instance, StreamingLLM~\cite{xiao2023attnsink} observes that attention scores frequently concentrate on the initial tokens and proposed retaining sink tokens during long-context modeling. H2O~\cite{zhang2023h2o} and its variants highlight the importance of intermediate tokens, utilizing token importance scores derived from attention maps to preserve pivotal tokens. There are also studies analyzing cache utilization across different heads in LLMs. For instance, DuoAttention~\cite{xiao2024duoattention} categorizes retrieval heads and streaming heads, compressing the KV cache of streaming heads to retain only recent tokens and attention sinks. FastGen~\cite{ge2023fastgen} further proposes a finer-grained head classification where each head handles tokens of varying lengths. However, these methods focus on token-level compression and do not consider frame-level redundancy in video data. Furthermore, we find that KV cache compression for video models can be more aggressive, with the cache of dummy heads potentially all removed.

\noindent
\textbf{Efficient Video Generation.}
Early video generation models typically adopt DiT~\cite{peebles2023dit} architectures with bidirectional attention. Since these methods generate all video frames in a single pass, they do not support KV caching. As a result, prior acceleration methods for video generation often rely on sparse attention to speed up fixed-length attention computation~\cite{zhang2025spargeattn,xia2025sparsevid,zhang2025silding_tile_attn,sun2025vorta}. For example, SVG~\cite{xi2025svg,yang2025svg2} introduces spatial and temporal heads to capture distinct intra-frame and inter-frame sparsity patterns. However, in autoregressive video diffusion models, attention mask shifts from bidirectional to causal, and the context length varies with autoregressive steps. This makes existing video sparse attention methods difficult to apply directly, as variable-length attention would require recompiling kernels at each step.

\section{Method}

\subsection{Preliminary} 
Autoregressive video diffusion models decompose video synthesis into a frame-by-frame process, where each autoregressive (AR) step generates a clean context via an iterative denoising process. Formally, given a video sequence consisting of 
$T$ frames $X = \{x_1, x_2, \cdots, x_T\}$, an autoregressive diffusion model is trained to learn the joint distribution under the following causal factorization: 
\begin{equation} 
p(x_1,x_2,\cdots,x_T)=\prod_{i=1}^{T}p(x_i \mid x_1,x_2,\cdots,x_{i-1}), 
\end{equation} where the conditional distribution $p(x_i | x_1,x_2,\cdots,x_{i - 1})$ is learned by denosing from a Gaussian distribution using a diffusion model~\cite{ho2020ddpm,song2020ddim,lipman2022flowmatching}. Notably, some works~\cite{huang2025selfforcing, yang2025longlive} model $x_i$ as a chunk consisting of multiple frames to encourage temporal consistency. For presentation clarity, we keep $x_i$ as a single-frame format in the following part of this paper.

Thanks to the autoregressive property, the condition from previous AR steps ($x_1,x_2,\cdots x_{i-1}$) can be modeled through the KV cache. To mitigate the quadratic computational complexity induced by increasing cache length, existing methods~\cite{yang2025longlive,krea_realtime_14b,liu2025rollingforcing} typically adopt a sliding-window strategy: a cache window of size $L$ is maintained, consisting of 1 sink frame and $L-1$ neighboring frames. In detail, the sink frame occupies only a small number of frames and remains fixed when the context window changes. For instance, in short video generation, most methods commonly select the first frame as the sink frame. The sink frame primarily serves as a global anchor via the attention sink mechanism~\cite{xiao2023attnsink,gu2024attentionsink2}, thereby encouraging temporal consistency across videos. In contrast, the neighbor frames are dynamically updated to capture temporal dependencies as the sliding window moves.

Under this cache management, the multi-head self-attention in existing autoregressive video diffusion models is trained to aggregate cross-frame information. Formally, denote the frame index of sink frame as $s$, at the $i$-th AR step, the self-attention layers work as follows: 
\begin{equation} 
\mathrm{softmax}(Q_i[K_s, K_{i-L+1:i-1},{K_i}]^\top)[V_s, V_{i-L+1:i-1},V_i], 
\end{equation}
where $[\cdot]$ denotes concatenation along token sequence.

\begin{figure}[!t]
    \centering
    \includegraphics[width=\linewidth]{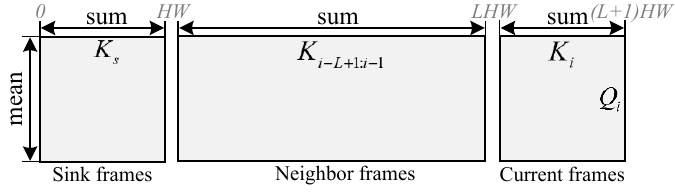}
    \vspace{-6mm}
    \caption{We compute the frame attention score by summing across rows and averaging across columns within the sink/neighbor/current frame group.}
    \label{fig:frame_attn_score}
    \vspace{-2mm}
\end{figure}

\subsection{Motivation}
\label{sec:motivation}

Current autoregressive video diffusions face efficiency challenges when handling long cache sequences. In this section, we delve into the multi-head self-attention layer to explore how the $i$-th frame interacts with its previous $L$ frames.

\noindent
\textbf{Profiling Setups.}
To quantitatively depict how queries from the current frame $Q_i$ interact with keys from the sink frames $K_s$, the neighbor frames $K_{i-L+1:i-1}$, and the current frame $K_i$, as shown in~\cref{fig:frame_attn_score}, we define the following ``frame attention score'':
\begin{equation}
\begin{aligned}
\label{eq:attn_cal}
&\alpha^{r} =\frac{1}{HW}\sum_{u=1}^{HW}\sum_{v\in\mathcal{J}_r}\mathcal{A}_{uv}, \\
&r \in \{\mathrm{sink, neighbor, current}\},
\end{aligned}
\end{equation}
where $\mathcal{A}\in \mathbb{R}^{HW \times (L+1)HW}$ denotes the attention map, $\mathcal{J}_\mathrm{sink}=[0,HW)$, $\mathcal{J}_{\mathrm{neighbor}}=[HW,LHW)$, $\mathcal{J}_\mathrm{current}=[LHW,(L+1)HW)$, and $HW$ represents the total number of visual tokens of one frame. Note that $\sum_r \alpha^r=1$. In the main paper, we use the Self Forcing model~\cite{huang2025selfforcing} as a representative, and provide results on other models in \suppl. Based on the above setup, we give the following three key observations.

\begin{figure*}[!t]
    \centering
    \includegraphics[width=0.99\linewidth]{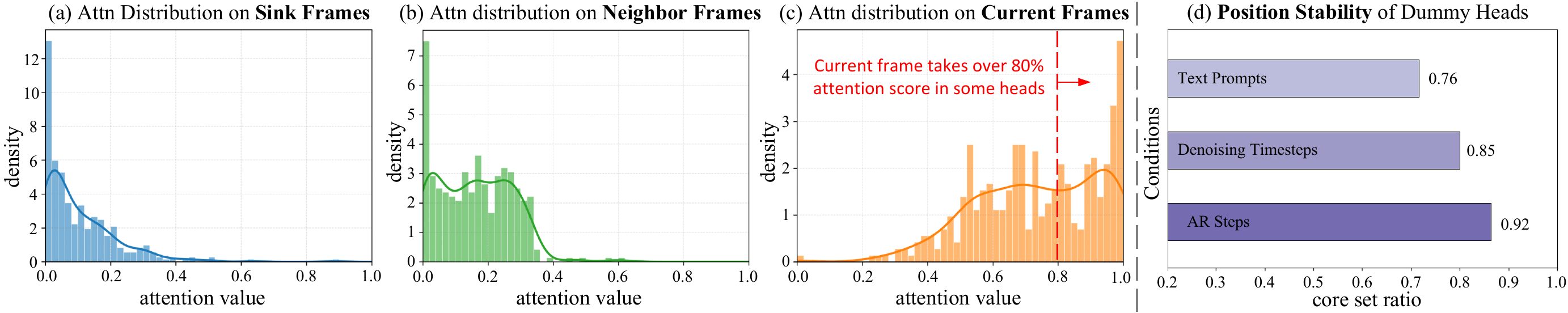}
    \vspace{-2mm}
    \caption{(a)-(c): We gather attention maps from all heads and use  ~\cref{eq:attn_cal} to compute the frame attention scores on the sink/neighbor/current frames. (d): The core set ratio under different conditions. For each bar, we change the corresponding condition while keeping the others fixed. We provide more implementation details of the observation experiment in the \suppl.}
    \label{fig:motivation}
    \vspace{-3mm}
\end{figure*}

\noindent
\textbf{\textit{Observation 1: Certain heads under-utilize past context.}} 
As shown in~\cref{fig:motivation}(c), we find there are about 25\% attention heads that assign over 80\% attention weights to the current frame. In other words, the $Q_i$ in these heads attends almost exclusively on $K_i$, even though $K_s$ and $K_{i-L+1:i-1}$ is available. Since these heads fail to work on frame aggregation, we refer to them as ``dummy heads'' in this paper. To identify a given number of $N$ dummy heads, we calculate the frame attention scores on current frames, i.e., $\alpha^{current}$, and then select the topN largest values as dummy heads.

\noindent
\textbf{\textit{Observation 2: Dummy head exhibits position stability.}}
We further study the dummy heads' position when conditions change, e.g., text prompts, AR steps, and denoising timesteps. Denote $\mathcal{I} = \{(l_n, h_n)\}_{n=1}^{N}$ as the dummy head location in the $l_n$-th layer and $h_n$-th head. We obtain $\mathcal{I}_c$, where $c=1,2,\cdots, C$, under $C$ varying conditions by computing $\alpha^{current}$ and select the topN. We then employ the core set ratio to quantify the variation of $\mathcal{I}_c$. The core set ratio is defined as  $\frac{1}{N}|\mathcal{I}_1\cap \mathcal{I}_2 \cap \cdots \cap \mathcal{I}_{C}| \in [0,1]$. As shown in~\cref{fig:motivation}(d), the position of dummy heads remained largely unchanged as condition changes. For example, 92\% head index repeatedly appears across all given AR steps.

\noindent
\textbf{\textit{Observation 3: Pruning cache of dummy heads incurs only slight drop.}}
Since little attention is paid to past frames in the dummy head, we attempt to remove their KV caches. Specifically, we sample one single condition and use TopN selection described above to drive $\mathcal{I}$ consisting of the location index of $N$ dummy heads, where $N$ is set to 25\% of the total number of heads. We then fix $\mathcal{I}$ and apply it throughout the evaluation on VBench~\cite{huang2024vbench}. We remove all KV caches for dummy heads while leaving non-dummy heads intact. As shown in~\cref{tab:motivation_obs3}, this naive compression strategy works with a 0.26\% drop. In contrast, random cache eviction severely degrades performance.

\begin{figure*}[!t]
    \centering
    \includegraphics[width=0.92\linewidth]{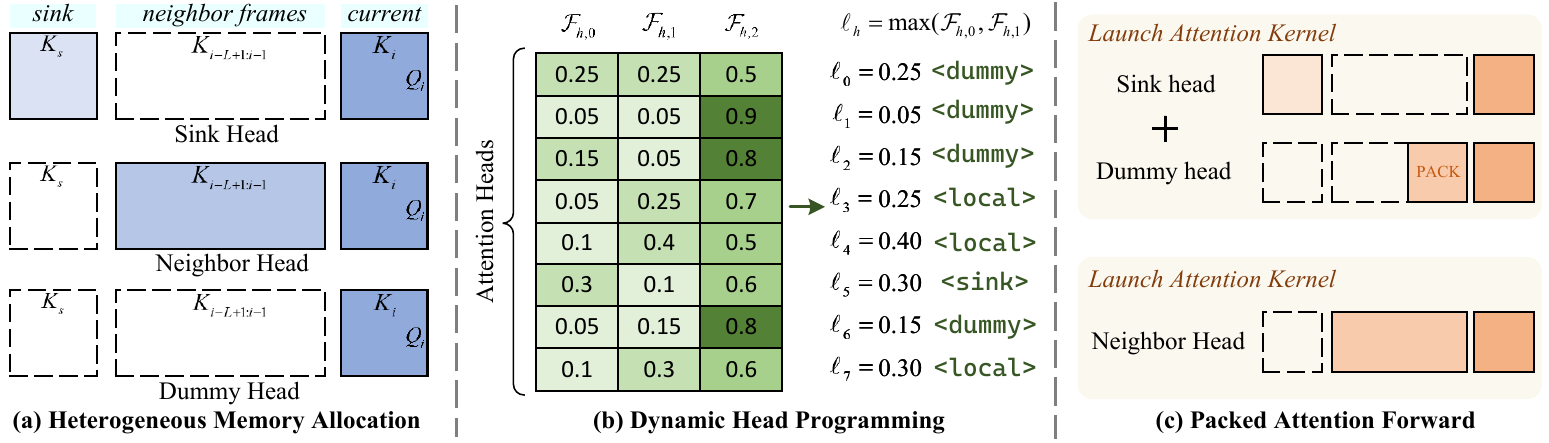}
    \vspace{-1mm}
    \caption{(a) We assign different contextual receptive fields to different head types. (b) A toy example of classifying different head types, with num\_head=8 and $N$=4 in this case. (c) We fuse different heads by context packing for more aggressive compression.}
    \label{fig:pipeline}
    \vspace{-3mm}
\end{figure*}

\subsection{Dummy Forcing}

Based on the observation in ~\cref{sec:motivation}, we propose \NAME to achieve more aggressive acceleration and better performance retention.

\begin{table}[!tb]
\centering
\caption{Results of pruning dummy heads' KV cache. ``Random'' denotes randomly selecting 25\% heads and evicting their caches.}
\label{tab:motivation_obs3}
\vspace{-2mm}
\setlength{\tabcolsep}{4pt}
\scalebox{0.9}{
\begin{tabular}{@{}lccccc@{}}
\toprule
setups       & FPS  & speedup     & Quality & Semantic & Total \\ \midrule
Self Forcing & 17.6 & 1.0$\times$ & 84.73   & 81.03    & 84.00 \\
Random       & 19.6 & 1.1$\times$ & 79.58   & 78.15    & 79.30 \\
Dummy Head   & 19.6 & 1.1$\times$ & 84.42   & 81.22    & 83.78 \\ \bottomrule
\end{tabular}%
}
\vspace{-3mm}
\end{table}

\noindent
\textbf{Overview.}
As shown in~\cref{fig:pipeline}, the proposed method comprises three components: (a) Heterogeneous Memory Allocation assigns head-specific KV cache lengths to achieve compression of redundant historical frames; (b) Dynamic Head Programming adaptively classify different head types by formalizing an optimization problem; (c) Packed Attention Forward extends context of dummy heads to mitigate the classification boundary between classes for aggressive dummy head numbers. More details are given below.

\noindent
\textbf{Heterogeneous Memory Allocation.}
Beyond compressing the context of the dummy head as shown in~\cref{sec:motivation}, we demonstrate in~\cref{sec:exp_ablation} that certain non-dummy heads exhibit high attention scores toward sink frames in the KV cache while utilizing few other historical frames. To further compress cache length, we subdivide non-dummy heads into ``sink heads'' and ``neighbor heads''. For sink heads, we remove redundant $L-1$ recent frames and restrict attention queries to only observe 1 sink frame and 1 current frame. For neighbor heads, we retain a sliding local window of size $L-1$ along with 1 current frame. Note that the sink frame is not included in the context of neighbor heads since it is exclusively modeled through sink heads. As shown in~\cref{fig:pipeline}(a), we categorize heads into three types: sink head, neighbor head, and dummy head, each possessing specific contextual information. Formally, at the $i$-th AR step, the self-attention is as follows:
\begin{equation}
\begin{aligned}
\label{eq:three_head_attn}
& \mathrm{sink}: \mathrm{softmax}(Q_i[K_{s},K_i]^{\top})[V_{s},V_i], \\
& \mathrm{neighbor}: \mathrm{softmax}(Q_iK_{i-L+1:i})V_{i-L+1:i},\\
& \mathrm{dummy}: \mathrm{softmax}(Q_iK^{\top}_i)V_i.
\end{aligned}
\end{equation}
Since attention heads are partitioned into three types, the original multi-head self-attention is modified as follows: we first extract three head groups through indexing along the head dimension. Then, we use ~\cref{eq:three_head_attn} to aggregate context according to different head types. Finally, we place the attention outputs back into their original positions. We use Triton~\cite{openai_triton} to reduce the additional overhead caused by the above head indexing and placement.

\noindent
\textbf{Dynamic Head Programming.}
As shown in ~\cref{fig:motivation}(d), the dummy head index $\mathcal{I}$ is not perfectly invariant under different conditions, e.g., there is about 25\% discrepancy when using varying text prompts. Intuitively, an adaptive head classification scheme would yield better performance. To achieve this, we first use ~\cref{eq:attn_cal} to compute each head's frame attention score $[\alpha^\mathrm{sink}, \alpha^\mathrm{neighbor}, \alpha^\mathrm{current}]\in \mathbb{R}^3$, which is then gathered across all heads to drive global frame attention score denoted as $\mathcal{F}\in\mathbb{R}^{\mathrm{num\_head }\times 3}$. Since $\mathcal{F}$ depends on the attention map, for computation efficiency, we uniformly sample a small portion of query tokens and compute an approximate attention map against all key tokens. We then formalize the adaptive head classification as the following optimization problem. Since different head classes incur corresponding frame attention loss, e.g., sink heads lack $\alpha^\mathrm{neighbor}$ due to the removal of neighbor frames, our objective is thus to maximize the remaining attention values while forcing $N$ dummy heads. Precisely, let $c_h \in \{\mathrm{sink}, \mathrm{neighbor}, \mathrm{dummy}\}$ be the head type at the $h$-th head, the optimization problem can be formularized as:
\begin{equation}
\begin{gathered}
\label{eq:objective}
\max \sum_{h=1}^{\mathrm{num\_head}} f_h(c_h), \\
\text{s.t. } \sum_{h=1}^{\mathrm{num\_head}} \mathbb{I}(c_h=\mathrm{dummy}) = N,
\end{gathered}
\end{equation}
where $f_h$ is the value function at the $h$-th attention head:
\begin{equation}
f_h(c_h) =
\left\{
\begin{aligned}
& \mathcal{F}_{h,0} + \mathcal{F}_{h,2} && c_h = \mathrm{sink}, \\
& \mathcal{F}_{h,1} + \mathcal{F}_{h,2} &&  c_h = \mathrm{neighbor}, \\
& \mathcal{F}_{h,2}           &&  c_h = \mathrm{dummy}.
\end{aligned}
\right.
\end{equation}

We point out that ~\cref{eq:objective} is a classic dynamic programming problem and proof in \suppl that the following greedy algorithm achieves optimality: for each head $h$, we compute $\ell_h = \max(\mathcal{F}_{h,0}, \mathcal{F}_{h,1})$ to obtain the opportunity cost of forcing the $h$-th head to be dummy. Then we sort $\{\ell_h\}_{h=1}^\mathrm{num\_head}$ and assign the $N$ smallest elements as dummy heads. For the remaining heads, we set $c_h=\mathrm{sink}$ if $ F_{h,0} \geq F_{h,1}$ else $c_h = \mathrm{neighbor}$.

\noindent
\textbf{Packed Attention Forward.}
As shown in~\cref{tab:motivation_obs3}, although removing KV caches of 25\% heads results in minimal performance loss, the resulting speedup remains limited. Intuitively, further increasing the number of dummy heads $N$ should bring additional acceleration. However, as shown in~\cref{sec:exp_ablation}, naively enlarging $N$ causes noticeable performance drops due to boundary mis-classification between non-dummy and dummy heads, pruning caches on context-critical heads. To address this, we propose to shift the classification boundary by slightly extending the context of dummy heads to additionally include some boundary heads. Specifically, given the temporal dependency decay in video frames, we additionally append the ($i-1$)-th frame as a packing frame into the context of dummy heads. After this improvement, the attention in dummy heads is as follows:
\begin{equation}
\mathrm{pack\_dummy}: \mathrm{softmax}(Q_i[K_{i-1},K_i]^{\top})[V_{i-1},V_{i}].
\end{equation}
As an additional benefit, dummy heads and sink heads now share the same effective context length. This enables us to pack both head types into a single attention call. As a result, the number of attention kernel launches is reduced from three to two. Empirically, the reduction in kernel launches largely compensates for the slightly enlarged context length, while enabling more than 50\% of heads to be configured as dummy heads without noticeable quality degradation.

\section{Experiments}

\subsection{Experimental Setup}

In the main paper, we apply the proposed method on state-of-the-art (SoTA) autoregressive diffusion methods, including Self Forcing~\cite{huang2025selfforcing} and LongLive~\cite{yang2025longlive}. We also provide results of other models in the \suppl. For tasks, we evaluate on classic video generation, including 5s short video and 30s long video in \cref{sec:classic_vid_gen}, high-resolution 720P\&1080P video generation in~\cref{sec:high_resolution_video_gen}, and long-context video generation in~\cref{sec:context_probing}.  Unless specified, we set the dummy head number to 50\% of total heads as the default. Due to page limit, we provide more implementation details in the \suppl.

\begin{table*}[!tb]
\centering
\caption{Quantitative comparison on efficiency and quality on \textbf{5 second short video generation} with VBench. The marks \textcolor{brickred}{$\circ$}, \textcolor{darkgreen}{$\diamond$},\textcolor{blueish}{$\bullet$} denote bidirectional models, autoregressive models, and inference acceleration methods, respectively. ``FPS'' denotes the number of frames generated per-second tested with one single H100 GPU. ``\textbf{\NAME}'' is our proposed method.}
\vspace{-2mm}
\label{tab:compare_short_vid_gen}
\setlength{\tabcolsep}{8pt}
\scalebox{0.88}{
\begin{tabular}{@{}lccccccc@{}}
\toprule
\multirow{2}{*}{Methods} & \multicolumn{4}{c}{\textbf{Inference Efficiency}}            & \multicolumn{3}{c}{\textbf{VBench Score}}                                     \\ \cmidrule(l){2-5}\cmidrule(l){6-8}
                         & \#Param & Resolution & FPS$\uparrow$ & Speedup$\uparrow$ & Quality Score$\uparrow$ & Semantic Score$\uparrow$ & Total Score $\uparrow$ \\ \midrule
\textcolor{brickred}{$\circ$}LTX-Video    & 1.9B & 768 $\times$ 512 & 8.98  & -             & 82.30 & 70.79 & 70.79 \\
\textcolor{brickred}{$\circ$}Wan2.1       & 1.3B & 832 $\times$ 480 & 0.78  & -             & 84.26 & 80.09 & 80.09 \\ 
\textcolor{darkgreen}{$\diamond$}SkyReels-V2  & 1.3B & 960 $\times$ 540 & 0.49  & -             & 82.67 & 84.70 & 74.53 \\
\textcolor{darkgreen}{$\diamond$}MAGI-1       & 4.5B & 832 $\times$ 480 & 0.19  & -             & 79.18 & 82.04 & 67.74 \\
\textcolor{darkgreen}{$\diamond$}CausVid      & 1.3B & 832 $\times$ 480 & 17.56 & -             & 81.20 & 84.05 & 69.80 \\ 
\textcolor{darkgreen}{$\diamond$}RealTime     & 14.0B & 832 $\times$ 480 & 2.89 & -             &85.24  &  81.69& 84.53 \\ \hdashline[0.5pt/2pt]
\textcolor{darkgreen}{$\diamond$}Self Forcing & 1.3B & 832 $\times$ 480 & 17.56 & \cellcolor{tab_green}1.0$\times$  & 84.73 & 81.03 & \cellcolor{tab_green}84.00 \\
\textcolor{blueish}{$\bullet$}+ Infinipot-V    & 1.3B & 832 $\times$ 480 & 19.54 &  \cellcolor{tab_green}1.1$\times$ & 82.72 & 79.05&\cellcolor{tab_green}81.99 \\
\textcolor{blueish}{$\bullet$}+ R-KV    & 1.3B & 832 $\times$ 480 & 18.41 &  \cellcolor{tab_green}1.1$\times$ &83.46 & 79.60 & \cellcolor{tab_green}82.69 \\
\textcolor{blueish}{$\bullet$}+ TeaCache    & 1.3B & 832 $\times$ 480 & 21.82 &  \cellcolor{tab_green}1.2$\times$ & 84.22 & 80.47 & \cellcolor{tab_green}83.47 \\
\textcolor{blueish}{$\bullet$}\textbf{+ \NAME}       & 1.3B & 832 $\times$ 480 & 24.30 & \cellcolor{tab_green}1.4$\times$ & 84.63 & 80.98 & \cellcolor{tab_green}83.90 \\ \hdashline[0.5pt/2pt]
\textcolor{darkgreen}{$\diamond$}LongLive     & 1.3B & 832 $\times$ 480 & 17.57 & \cellcolor{tab_blue}1.0$\times$  & 83.64 & 81.08 & \cellcolor{tab_blue}83.13 \\
\textcolor{blueish}{$\bullet$}+ Infinipot-V    & 1.3B & 832 $\times$ 480 & 19.54 &  \cellcolor{tab_blue}1.1$\times$ & 82.70 &80.92 &\cellcolor{tab_blue}82.35 \\
\textcolor{blueish}{$\bullet$}+ R-KV    & 1.3B & 832 $\times$ 480 & 18.41 &  \cellcolor{tab_blue}1.1$\times$ & 82.92 &81.26 &\cellcolor{tab_blue}82.59 \\
\textcolor{blueish}{$\bullet$}+ TeaCache    & 1.3B & 832 $\times$ 480 & 21.82 &  \cellcolor{tab_blue}1.2$\times$ & 83.10 & 80.73 & \cellcolor{tab_blue}82.63\\
\textcolor{blueish}{$\bullet$}\textbf{+ \NAME}   & 1.3B & 832 $\times$ 480 & 24.30 & \cellcolor{tab_blue}1.4$\times$  & 83.28 & 80.89 & \cellcolor{tab_blue}82.80 \\ \bottomrule
\end{tabular}%
}
\vspace{-2mm}
\end{table*}

\begin{table}[!tb]
\centering
\caption{Quantitative comparison on efficiency and quality on \textbf{30s long video generation} with VBench-Long. }
\label{tab:compare_long_vid_gen}
\vspace{-2mm}
\setlength{\tabcolsep}{4pt}
\scalebox{0.88}{
\begin{tabular}{@{}lccccc@{}}
\toprule
method & FPS$\uparrow$ & Speedup$\uparrow$ & Quality$\uparrow$ & Semantic$\uparrow$ & Total$\uparrow$ \\ \midrule
SkyReels-V2 & 0.49 & - & 80.77 & 53.37 & 75.29 \\
FramePack & 0.92 & - & 83.61 & 75.32 & 81.95 \\ \hdashline[0.5pt/2pt]
Self-Forcing & 17.56 & \cellcolor{tab_green} 1.0$\times$ & 84.36 & 80.18 &  \cellcolor{tab_green}83.53 \\
+TeaCache & 21.82 &\cellcolor{tab_green}1.2$\times$ & 83.82  & 79.74 & \cellcolor{tab_green}83.00 \\
\textbf{+Ours} & 24.30 &  \cellcolor{tab_green}1.4$\times$ & 84.28 & 78.85 &  \cellcolor{tab_green}83.19 \\ \hdashline[0.5pt/2pt]
LongLive & 17.57 &  \cellcolor{tab_blue}1.0$\times$ & 83.34 & 80.34 & \cellcolor{tab_blue}82.74 \\
+TeaCache  & 21.82 &\cellcolor{tab_blue}1.2$\times$ & 82.87 & 80.20 & \cellcolor{tab_blue}82.33 \\
\textbf{+Ours} & 24.30 &\cellcolor{tab_blue}1.4$\times$ &83.33  & 79.56 & \cellcolor{tab_blue}82.57 \\ \bottomrule
\end{tabular}%
}
\end{table}

\subsection{Classic Video Generation}
\label{sec:classic_vid_gen}

\noindent
\textbf{Comparison on Short Video Generation.}
Since few studies have explored accelerating autoregressive video diffusion models, we carefully select SoTA methods from related domains for comparison with other inference speedup approaches. Specifically, R-KV~\cite{cai2025rkv} employs token-level KV cache compression for LLM reasoning, Infinipot-V~\cite{kim2025infinipot} prunes caches for streaming video understanding, and TeaCache~\cite{liu2024teacache} skips denoising timesteps for DiT acceleration. We report results in ~\cref{tab:compare_short_vid_gen}. For previous KV cache pruning methods, R-KV and Infinipot-V, since they calculate token importance at each AR step, the additional time introduced by token selection algorithm undermines the benefits gained from reduced cache length, resulting in a marginal 1.1$\times$ overall speedup ratio. For diffusion step skipping method TeaCache, the acceleration gain is limited given that current base models are already few-step diffusion models. In contrast, our method achieves the most aggressive acceleration while maintaining the best performance. For instance, in Self Forcing, our \NAME achieves 1.4$\times$ speedup, generating video at 24.3FPS in real time with only a 0.1\% drop in quality. In \suppl, we provide further discussion on cache compression ratios, speedup effects on a single attention layer and qualitative visualization results.

\begin{figure*}[!t]
    \centering
    \includegraphics[width=0.98\linewidth]{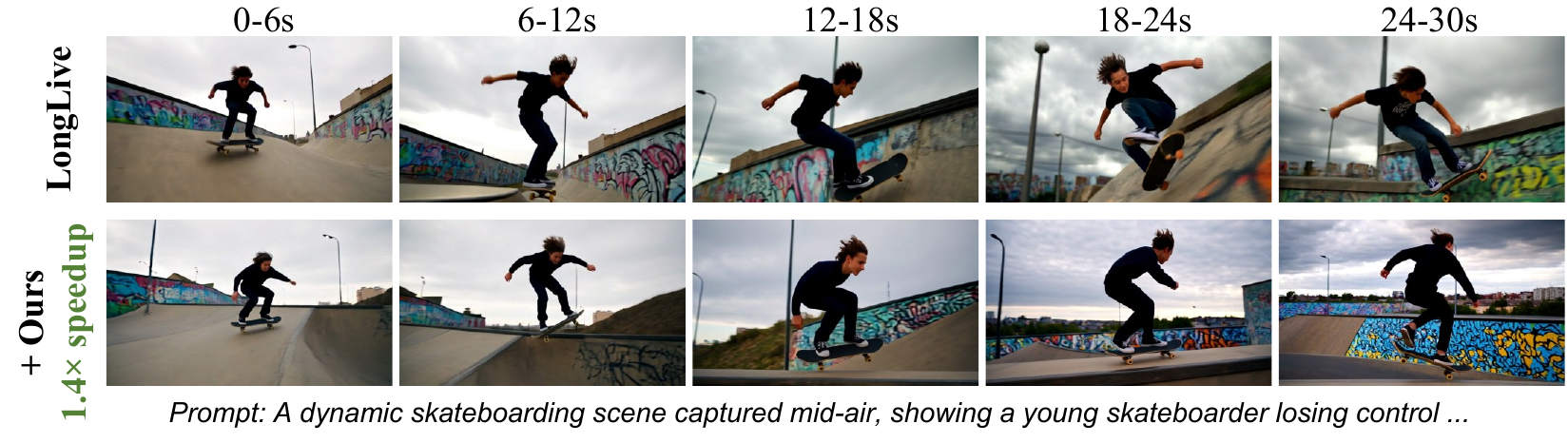}
    \vspace{-3mm}
    \caption{Quantitative comparison on  \textbf{30s long video generation }. Our \NAME achieves $1.4\times$ end-to-end acceleration without compromising output quality. More results are provided in the \suppl.}
    \label{fig:main_single30s_longlive}
    \vspace{-2mm}
\end{figure*}

\noindent
\textbf{Comparison on Long Video Generation.}
~\cref{tab:compare_long_vid_gen} gives the VBench-Long~\cite{huang2024vbench++} evaluation results on generating 30s long videos. Our \NAME achieves better efficiency and performance than competitive benchmarks, allowing 1.4$\times$ end-to-end speedup with up to a 0.4\% quality drop. In~\cref{fig:main_single30s_longlive}, we present quantitative comparison results, and our method achieves faster generation while maintaining strong motion and high visual quality.

\begin{table}[!]
\centering
\caption{Quantitative VBench results on \textbf{high-resolution video generation} at 720P (1280$\times$720) and 1080P (1920$\times$1088).}
\label{tab:high_resolution_vid_gen}
\vspace{-2mm}
\setlength{\tabcolsep}{1pt}
\scalebox{0.88}{
\begin{tabular}{@{}lcccccc@{}}
\toprule
Methods & resolution & FPS$\uparrow$ & speedup$\uparrow$ & Quality$\uparrow$ & Semantic$\uparrow$ & Total$\uparrow$ \\ \midrule
Self Forcing & 720P & 5.6 & \cellcolor{tab_green}1.0$\times$ & 84.77 & 81.93 & \cellcolor{tab_green}84.20 \\
\textbf{+Ours} & 720P & 9.1 & \cellcolor{tab_green}1.6$\times$ & 84.77 & 81.61 & \cellcolor{tab_green}84.14 \\
LongLive & 720P & 5.6 & \cellcolor{tab_blue}1.0$\times$ & 83.82 & 81.50 & \cellcolor{tab_blue}83.36 \\
\textbf{+Ours} & 720P & 9.1 & \cellcolor{tab_blue}1.6$\times$ & 83.21 & 81.91 & \cellcolor{tab_blue}82.95 \\ \hdashline[0.5pt/2pt]
Self Forcing & 1080P & 1.3 & \cellcolor{tab_green}1.0$\times$ & 84.90 & 78.68 & \cellcolor{tab_green}83.66 \\
\textbf{+Ours} & 1080P & 2.6 & \cellcolor{tab_green}2.0$\times$ & 84.15 & 78.76 & \cellcolor{tab_green}83.07 \\
LongLive & 1080P & 1.3 & \cellcolor{tab_blue}1.0$\times$ & 81.86 & 80.78 & \cellcolor{tab_blue}81.65 \\
\textbf{+Ours} & 1080P & 2.6 & \cellcolor{tab_blue}2.0$\times$ & 81.97 & 80.39 & \cellcolor{tab_blue}81.65 \\ \bottomrule
\end{tabular}%
}
\end{table}

\subsection{High-resolution Video Generation.}
\label{sec:high_resolution_video_gen}
As resolution increases, the number of cached visual tokens grows quadratically, leading to inefficiency in existing methods. In this section, we explore applying the proposed \NAME in generating videos at higher resolutions such as 720P and 1080P. Specifically, we discover that current autoregressive video diffusion models exhibit strong zero-shot capabilities for low-to-high resolution video generation. To leverage this, we modify the shape of the initial Gaussian noise in each AR step to allow high-resolution generation. Experimental results are presented in~\cref{tab:high_resolution_vid_gen}. As the length of tokens to be processed increases, the speedup achieved by our method becomes more pronounced. For instance, ours achieves even 2.0$\times$ acceleration without quality drop on 1080P video with LongLive model. This result demonstrates the generalizability of the proposed method.

\begin{figure*}[!t]
    \centering
    \includegraphics[width=0.98\linewidth]{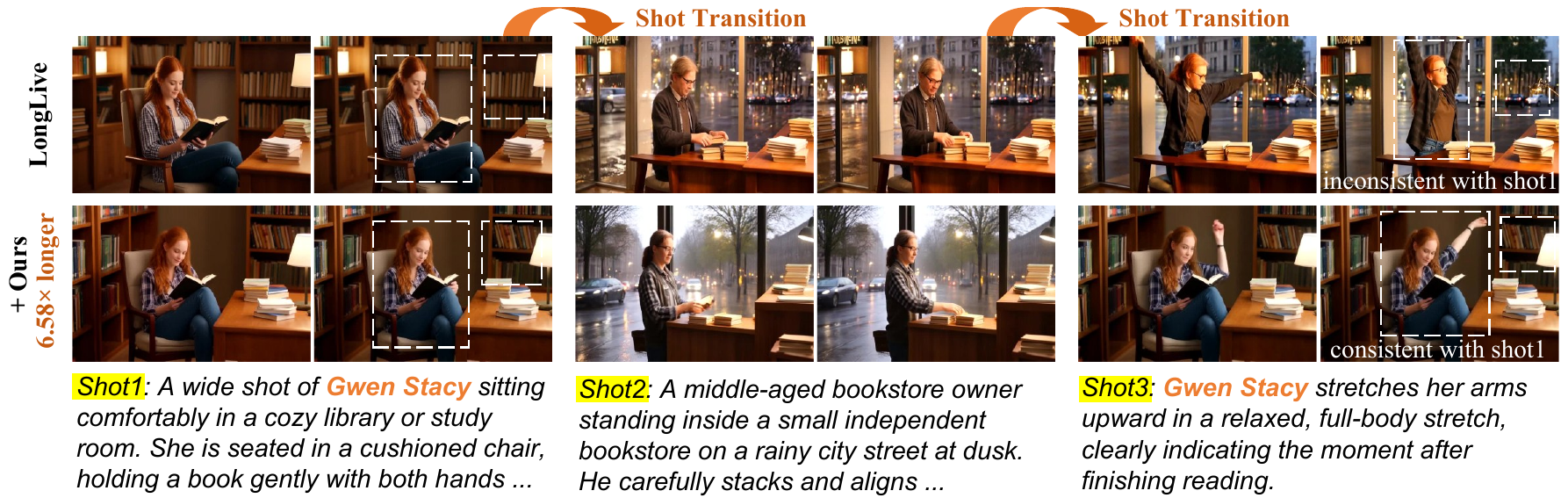}
    \vspace{-1mm}
    \caption{Quantitative comparison on \textbf{long-context video generation}. We design a set of text prompts with a storytelling narrative that involves two scene transitions, where each shot is interactively generated through the corresponding prompt. See more results in~\suppl.}
    \vspace{-3mm}
    \label{fig:main_context_probing}
\end{figure*}

\subsection{Long-context Video Generation}
\label{sec:context_probing}

Current autoregressive video generation models typically employ sliding windows to avoid excessively long sequences. For instance, SoTA methods~\cite{yang2025longlive} can only observe the past 36 frames. Consequently, during shot transitions in narrative scenarios, e.g., a character disappears and reappears, current methods inevitably produce inconsistent videos. Here, we explore long-context video generation to evaluates different methods' historical context length under similar runtime cost. Detailed experimental settings are given in \suppl. For our \NAME, we allocate cache budget saved from the dummy\&sink heads to the neighbor heads, thereby enabling longer effective cache. For comparative baselines, we consider the LongLive model with and without the sliding window strategy. As shown in~\cref{tab:compare_long_context}, LongLive without a local window suffers from computational complexity when processing long visual tokens. In contrast, the proposed \NAME can achieve similar cache length while delivering a 1.93$\times$ speedup. On the other hand, compared with LongLive using a sliding window, our \NAME achieves better generation quality (68.45 v.s. 69.48) thanks to our 6.58$\times$ longer cache size. Furthermore, we provide a quantitative comparison in~\cref{fig:main_context_probing}. As can be seen, current methods tend to ``re-generate'' a new identity due to limited context, while ours can accurately reproduce previously vanished characters and backgrounds. At last, we also evaluate on 60s interactive video generation, see the \suppl for results.

\begin{table}[!tb]
\centering
\caption{Quantitative comparison on \textbf{long-context video generation}. \#cache denotes the number of past cached frames. ``sw'' means using the local sliding window strategy.}
\vspace{-2mm}
\label{tab:compare_long_context}
\setlength{\tabcolsep}{4pt}
\scalebox{0.88}{
\begin{tabular}{@{}lccccc@{}}
\toprule
\multirow{2}{*}{setups} & \multicolumn{2}{c}{\textbf{Context}} & \multicolumn{2}{c}{\textbf{Efficiency}} & \multirow{2}{*}{\textbf{Score$\uparrow$}} \\ \cmidrule(l){2-3} \cmidrule(l){4-5}
 & \#cache$\uparrow$ & ctx\_ratio$\uparrow$ & FPS$\uparrow$ & speedup$\uparrow$ &  \\ \midrule
LongLive w/o sw & 237 & 6.58$\times$ & 9.36 &  \cellcolor{tab_blue}1.00$\times$ & 69.38 \\
LongLive w/ sw & 36 &  \cellcolor{tab_green}1.00$\times$ & 17.57 & 1.87$\times$ & 68.45 \\
\textbf{\NAME} & 237 &  \cellcolor{tab_green}6.58$\times$ & 18.14 &  \cellcolor{tab_blue}1.93$\times$ & 69.48 \\ \bottomrule
\end{tabular}%
}
\end{table}

\subsection{Ablation Studies}
\label{sec:exp_ablation}

\begin{table}[!tb]
\centering
\caption{Ablation experiments of different components on VBench.}
\label{tab:ablation_on_components}
\vspace{-2mm}
\setlength{\tabcolsep}{3pt}
\scalebox{0.88}{
\begin{tabular}{@{}llccc@{}}
\toprule
setups & FPS & Quality & Semantic & Total \\ \midrule
(0)Self Forcing (baseline) & 17.6 & 84.73 & 81.03 & 84.00 \\
(1)combine sink\&neighbor & 22.1 & 84.23 & 81.09 & 83.60 \\
(2)w/o packing in dummy head & 24.2 & 83.91 & 80.86 & 83.30 \\
(3)\NAME (ours) & 24.3 & 84.63 & 80.98 & 83.90 \\ \bottomrule
\end{tabular}%
}
\end{table}

\noindent
\textbf{Effectiveness of Different Components.}
To verify different design choices in the proposed method, we ablate on other alternative configurations in~\cref{tab:ablation_on_components}.
(1) We combine sink and neighbor head class into a unified non-dummy head type, whose context consists of 1 sink frame and $L-1$ recent frames. Results show this variant achieves reasonable performance, but its acceleration is suboptimal. This is because although both sink head and neighbor head are non-dummy heads, they focus on distinct parts of the context, and simply merging them would result in redundancy. (2) We remove the packing frame in dummy heads and call three separate attention. Results show that it degrades performance while offering negligible speed gains. This is because naively increasing dummy head number without adjusting the classification boundary between dummy and non-dummy impairs the information aggregation of context-critical heads. And the runtime from additional attention calls negates the benefits from reduced cache length.

\begin{figure}[!t]
    \centering
    \includegraphics[width=\linewidth]{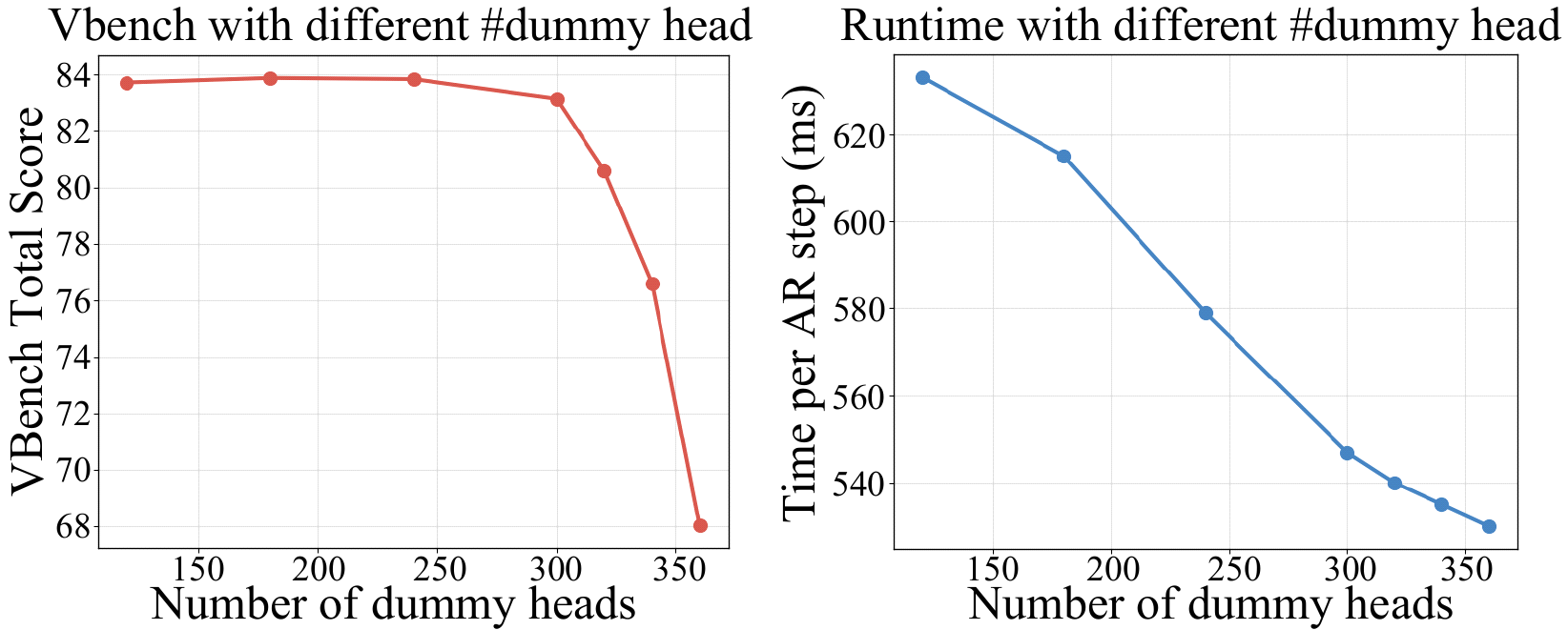}
    \vspace{-5mm}
    \caption{Ablation experiments on Vbench scores and average runtime per AR step across different numbers of dummy heads.}
    \label{fig:ablation_dummy_head_ratio}
    \vspace{-3mm}
\end{figure}

\noindent
\textbf{Different Dummy Ratios.}
In the proposed \NAME, the number of dummy heads $N$ serves as a critical hyperparameter to balance efficiency and performance. Here, we set different $N$ to investigate its impact. We employed the Self Forcing model, which consists of a total of 360 heads. As shown in ~\cref{fig:ablation_dummy_head_ratio}, when $N$ is set below 240, the proposed method maintains relatively stable performance compared to $N=0$. This phenomenon indicates that nearly 2/3 heads do not fully utilize past frames. In contrast, reducing the dummy head count from 300 to 360 causes significant degradation. In this case, KV cache pruning harms neighbor heads, which are crucial for context aggregation.

\section{Discussion}

\noindent
\textbf{Compatibility with other Acceleration Methods.}
Since our \NAME works across attention heads, we further combine it with the previously best-performing method TeaCache, yielding a variant ``Ours+TeaCache''. As shown in~\cref{tab:compatible_with_teacache}, the combination of these two orthogonal methods achieves larger generation speeds, e.g., over 30FPS, demonstrating the potential for integration with other methods.

\begin{figure}[!t]
    \centering
    \includegraphics[width=0.99\linewidth]{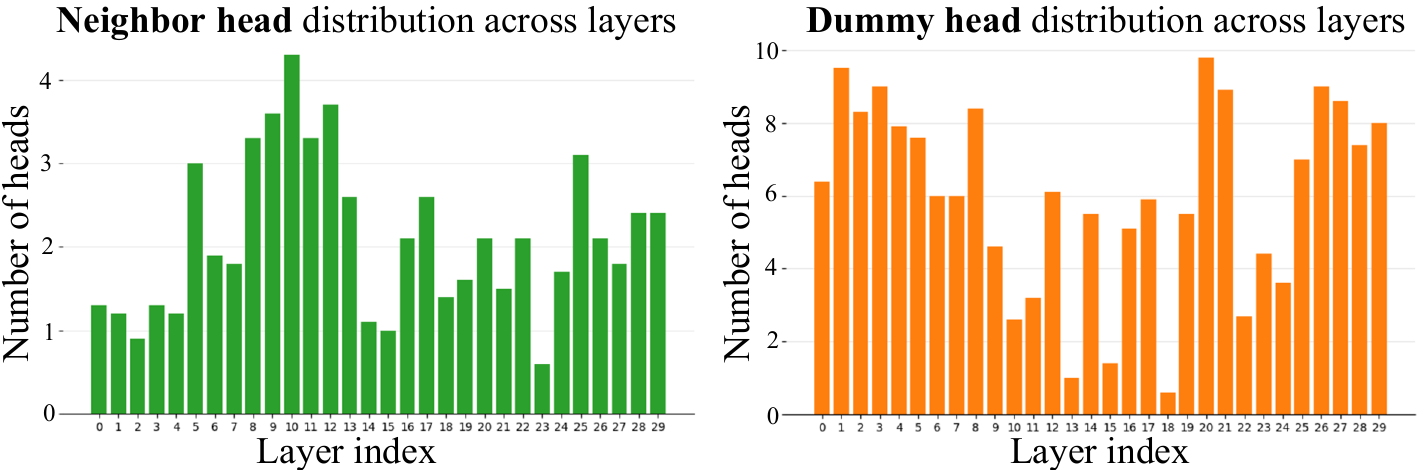}
    \vspace{-2mm}
    \caption{Distribution of the number of neighbor heads and dummy heads across different layers.}
    \label{fig:ablation_dummy_head_dist}
\end{figure}

\noindent
\textbf{Head Distribution across Layers.}
Since the proposed method can automatically identify dummy head positions, it is interesting to observe the distribution of different head classes across layers. In ~\cref{fig:ablation_dummy_head_dist}, we present the number of neighbor and dummy heads per layer averaged across 100 prompts. Roughly, dummy heads primarily appear in the first and last few layers, while neighbor heads cluster in intermediate layers. To explain this, the model first aggregates information from the current frame in shallow layers to abstract high-level features. Subsequently, it queries past frames in this high-level semantic space in intermediate layers for information aggregation. In the last few layers, the model refines the current frame and returns low-level space for subsequent decoding.

\begin{table}[!tb]
\centering
\caption{Results of combining our method and TeaCache.}
\label{tab:compatible_with_teacache}
\vspace{-2mm}
\setlength{\tabcolsep}{1.5pt}
\scalebox{0.88}{
\begin{tabular}{@{}lcccccc@{}}
\toprule
\multirow{2}{*}{methods} & \multicolumn{3}{c}{\textbf{Self-Forcing}} & \multicolumn{3}{c}{\textbf{LongLive}} \\ \cmidrule(l){2-4} \cmidrule(l){5-7}
                & FPS  & speedup     & VBench & FPS  & speedup     & VBench \\ \midrule
original models & 17.6 & 1.0$\times$ & 84.00  & 17.6 & 1.0$\times$ & 83.13  \\
+Ours\&TeaCache & 30.5 & 1.7$\times$ & 83.26  & 30.5 & 1.7$\times$ & 82.47  \\ \bottomrule
\end{tabular}%
}
\vspace{-3mm}
\end{table}

\section{Conclusion}
In this work, we identify the dummy heads in existing autoregressive video diffusion models. Building on this observation, we propose \NAME, which employs heterogeneous memory allocation to expose different attention heads with varying context lengths. We further introduce dynamic head programming that derives an optimal greedy policy to perform online head classification. For aggressive dummy head numbers, we propose packed attention forward, which extends context length without incurring overhead. We apply \NAME to multiple downstream tasks, including efficient video generation, high-resolution video generation, and long-context video generation. Extensive experiments demonstrate the generality of our method.

\section*{Impact Statement}
The primary goal of this work is to advance the efficiency and scalability of machine learning models, enabling faster video generation and more effective use of computational resources without additional training. Since this work is developed based on pre-trained video generation models, potential risks associated with synthetic video generation, such as misinformation or misuse, are inherited from the broader class of generative video models.  We expect the broader impact of this work to be largely positive by facilitating more efficient and scalable deployment of advanced video generation models while remaining aligned with existing ethical considerations in generative modeling.

\bibliography{icml}
\bibliographystyle{icml2026}

\clearpage
\newpage
\appendix
\onecolumn

\section{Proof of Optimality for Greedy Strategy}
\label{sec:suppl-proof-optimal}
In the proposed Dynamic Head Programming, we demonstrate that the dummy head assignment is in fact a dynamic programming problem that can be efficiently solved with $\mathcal{O}(n\log n)$ complexity using a greedy algorithm. This section provides a proof of optimality. 

Recall that the original objective is 
\begin{equation}
\begin{gathered}
\max \sum_{h=1}^{\mathrm{num\_head}} f_h(c_h), \\
\text{s.t. } \sum_{h=1}^{\mathrm{num\_head}} \mathbb{I}(c_h=\mathrm{dummy}) = N,
\end{gathered}
\end{equation}
where $f_h$ is the value function at the $h$-th attention head:
\begin{equation}
f_h(c_h) =
\left\{
\begin{aligned}
& \mathcal{F}_{h,0} + \mathcal{F}_{h,2} && c_h = \mathrm{sink}, \\
& \mathcal{F}_{h,1} + \mathcal{F}_{h,2} &&  c_h = \mathrm{neighbor}, \\
& \mathcal{F}_{h,2}           &&  c_h = \mathrm{dummy}.
\end{aligned}
\right.
\end{equation}

Denote $\mathcal{I}^{*}$ as the set of dummy head indices obtained from our greedy strategy. This solution achieves a total value of 
\begin{equation}
V^{*} = \sum_{h \notin \mathcal{I}^{*}} \max(\mathcal{F}_{h,0}, \mathcal{F}_{h,1}) + \sum_{h=1}^{\mathrm{num\_head}} \mathcal{F}_{h,2}.
\end{equation}

Consider any alternative dummy head assignment solution $\mathcal{I}\neq \mathcal{I}^*$, where $|\mathcal{I}|=N$. Since $\mathcal{I}\neq \mathcal{I}^*$, there exist head index $i\in \mathcal{I} /\ \mathcal{I}^*$ and $j \in \mathcal{I}^* /\ \mathcal{I}$, i.e., $i$-th head is selected as dummy head in $\mathcal{I}$ but not in $\mathcal{I}^*$, and vice versa for $j$-th head. According to the definition of our greedy algorithm, we can derive $\ell_i \geq \ell_j$, i.e., 
\begin{equation}
\max(\mathcal{F}_{i,0},\mathcal{F}_{i,1}) \geq \max(\mathcal{F}_{j,0},\mathcal{F}_{j,1}).
\end{equation}

The value of a given policy $\mathcal{I}$ is  $V_{\mathcal{I}} =   \sum_{h \notin \mathcal{I}} \max(\mathcal{F}_{h,0}, \mathcal{F}_{h,1}) + \sum_{h=0}^\mathrm{num\_head} \mathcal{F}_{h,2}.$ Consider a swap operation on $\mathcal{I}$ which excludes $i$-th head and  includes $j$-th head, i.e. $\mathcal{I}^\mathrm{swap} = \mathcal{I} \setminus \{i\} \cup \{j\}$.  Then the total value change due to this exchange is:
\begin{equation}
V_{\mathcal{I}^\mathrm{swap}} - V_{\mathcal{I}} = \max(\mathcal{F}_{i,0}, \mathcal{F}_{i,1}) - \max(\mathcal{F}_{j,0}, \mathcal{F}_{j,1})
= \ell_i - \ell_j \ge 0.
\end{equation}

Therefore, $V_{\mathcal{I}^\mathrm{swap}} \geq V_{\mathcal{I}}$. By repeatedly applying such a swap operation, we can transform any solution $\mathcal{I}$ into our greedy solution $\mathcal{I}^*$ without decreasing the total objective value. Since each exchange either strictly increases the value or leaves it unchanged, and the process terminates at $\mathcal{I}^*$, the greedy algorithm is thus optimal. \hfill $\square$

\section{Dummy Head in Other Models}
\label{sec:suppl_general_dummy_head}

In the main paper, we present the dummy head observation based on the Self Forcing~\cite{huang2025selfforcing} model. In this section, we extend to other popular autoregressive video diffusion frameworks, including CausVid~\cite{chen2024diffusionforcing} and Rolling Forcing~\cite{liu2025rollingforcing}. Results are as follows.

\begin{figure*}[!t]
    \centering
    \includegraphics[width=0.98\linewidth]{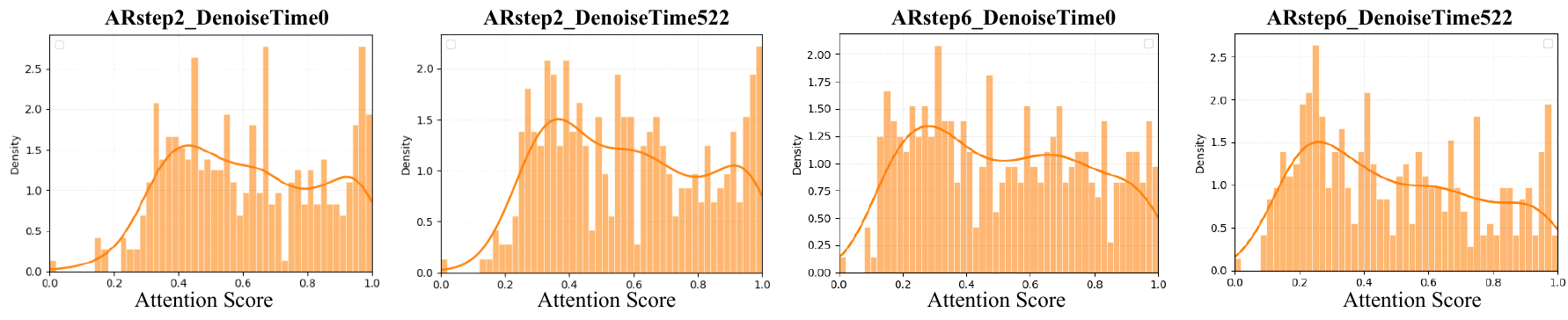}
    \vspace{-3mm}
    \caption{Frame attention scores on the current frame in the CausVid model~\cite{yin2025causvid}. We present the distribution across different AR steps and denoising timesteps.}
    \label{fig:suppl_CausVid_dist}
\end{figure*}

CausVid~\cite{yin2025causvid} trains video models employing the Diffusion Forcing~\cite{chen2024diffusionforcing} framework, which conditions the current frame on previous frames with varying noise levels. Under this paradigm, the self-attention layer has to additionally include the noise distribution discrepancy across frames. Similar to~\cref{sec:motivation}, we present the frame attention score on the current frame in~\cref{fig:suppl_CausVid_dist}. It can be observed that under different conditions, there are heads exhibiting attention scores larger than 0.8 on the current frame, indicating the existence of dummy heads. For further validation, we apply the proposed \NAME to this model by setting 50\% heads as dummy heads. The generated videos are shown in~\cref{fig:suppl_CausVid_visual}. It can be seen that removing 50\% of the total KV cache does not cause significant quality degradation, demonstrating the applicability of dummy heads within the Diffusion Forcing framework.

\begin{figure*}[!t]
    \centering
    \includegraphics[width=0.98\linewidth]{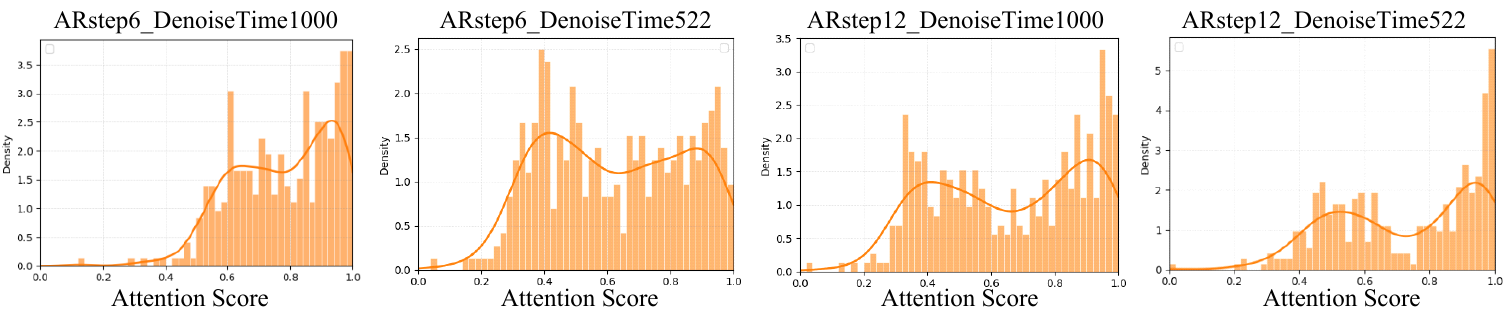}
    \vspace{-3mm}
    \caption{Frame attention scores on the current frame in the Rolling Forcing model~\cite{liu2025rollingforcing}. We present the distribution across different AR steps and denoising timesteps.}
    \label{fig:suppl_RollingForcing_dist}
\end{figure*}

Rolling Forcing~\cite{liu2025rollingforcing} proposes a joint denoising scheme that simultaneously denoises multiple frames within a rolling window, where each frame exhibits progressively increasing noise levels. During each model forward, Rolling Forcing performs denoising on frames within the window using conditions from previous clean frames. In~\cref{fig:suppl_RollingForcing_dist}, we present the frame attention distribution on the current frame, and one can see that the dummy head also exists. We then use \NAME on this model with 50\% dummy heads and provide the generated results in~\cref{fig:suppl_rollingForcing_visual}. As can be seen, there is no noticeable degradation in video quality, demonstrating the versatility of the dummy head in the Rolling Forcing framework.

\section{More Implementation Details}
\label{sec:suppl_more_impl_details}

\noindent
\textbf{Setup for Observation Experiments.}
In this section, we provide detailed settings for the profiling experiments described in~\cref{sec:motivation}. Specifically, for~\cref{fig:motivation}(a)-(c), we collect attention maps across all attention heads of the model. We use the average results from 100 text prompts sampled from VBench. Additionally, we select the third AR step because the lengths of attention key for the sink/neighbor/current frames are identical at this step. We employ the last denoising step to compute attention maps to avoid the impact from noise. For~\cref{fig:motivation}(d), given the core-set ratio $\frac{1}{N}|\mathcal{I}_1\cap \mathcal{I}_2 \cap \cdots \cap \mathcal{I}_{C}|$ monotonically decreases with increasing $C$, we set $C=5$ since existing models typically perform diffusion denoising for up to five steps. In this way, conditions including text prompts, AR steps and denoising time are controlled to exhibit the same value scales. For the results in~\cref{tab:motivation_obs3}, we employ a single text prompt `` a bicycle accelerating to gain speed'' along with the third AR step and the last denoising step to obtain the attention map for all heads. Based on this attention map, we select the TopN largest $\alpha^{current}$ values to determine the fixed dummy head positions during Vbench evaluations.

\noindent
\textbf{Setup for Dummy Forcing Implementation.}
Here, we provide the specific implementation details of the proposed \NAME algorithm. Specifically, since the frame attention score $\mathcal{F}$ in our method requires attention maps, we uniformly sampled 25\% of the original query tokens for efficient estimation. These tokens were then used to compute attention maps with all key tokens. We find this approximation achieves good performance while taking less than 10ms in practice. Additionally, the proposed method relies on the head classification results for head-specific KV cache management. To achieve this, we select the third AR step and the last denoising timestep across all attention heads to classify head types. Thanks to the stability of the dummy head as demonstrated in ~\cref{sec:motivation}, we then keep this head classification result fixed during subsequent AR steps and denoising time steps. In other words, our method requires only a single head classification computation call which can complete within 100ms. In contrast, previous KV cache methods require recalculation at each AR steps to update the KV cache tokens, incurring additional computational overhead.

\section{More Experimental Results}
\label{sec:suppl_more_exp_results}

\begin{table*}[!tb]
\centering
\caption{Results on 60s interactive long video generation. Quality scores are reported based on the whole video. CLIP semantic scores are reported on every 10-second video clip.}
\label{tab:compare_interactive_long_vid_gen}
\setlength{\tabcolsep}{6pt}
\scalebox{0.88}{
\begin{tabular}{@{}lcccccccccc@{}}
\toprule
\multirow{2}{*}{Methods} & \multirow{2}{*}{FPS$\uparrow$} & \multirow{2}{*}{Speedup$\uparrow$} & \multirow{2}{*}{Quality Score$\uparrow$} & \multicolumn{7}{c}{CLIP Semantic Score$\uparrow$} \\ \cmidrule(l){5-11} 
 &  &  &  & 0–10s & 10–20s & 20–30s & 30–40s & 40–50s & 50–60s & Avg. \\ \midrule
Self-Forcing & 17.56 & 1.0$\times$ & 82.77 & 0.3431 & 0.3353 & 0.3303 & 0.3245 & 0.3175 & 0.3219 & 0.3288 \\
\textbf{+Dummy Forcing} & 24.30  & 1.4$\times$ & 83.38 & 0.3389 & 0.3266 & 0.3157 & 0.3226 & 0.3108 & 0.3193 & 0.3223 \\
\hdashline[0.5pt/2pt]
LongLive & 17.57 & 1.0$\times$ & 84.63 & 0.3460 & 0.3370 & 0.3355 & 0.3298 & 0.3216 & 0.3223 & 0.3320 \\
\textbf{+Dummy Forcing} & 24.30 & 1.4$\times$ & 84.38 & 0.3367 & 0.3264 & 0.3300 & 0.3197 & 0.3119 & 0.3260 & 0.3251 \\ \bottomrule
\end{tabular}%
}
\end{table*}

\noindent
\textbf{Comparison on 60s Interaction Video Generation.}
Following ~\citet{yang2025longlive}, we generate a 60-second-long video by sequentially feeding the model with 6 text prompts, each responsible for a 10-second video segment. Since VBench does not include corresponding tasks, we adopt the 12 prompt suites released by LongLive. For evaluation, we assess the generated videos on VBench-Long using dimensions that support customized-prompt videos, including subject consistency, background consistency, motion smoothness, aesthetic quality, and imaging quality. Moreover, we compute the CLIP~\cite{radford2021CLIP} similarity between each video segment and its corresponding text prompt to evaluate semantic adherence. We employ the KV re-caching~\cite{yang2025longlive} technique to mitigate temporal shift in long video generation. The experimental results are summarized in~\cref{tab:compare_interactive_long_vid_gen}. Our method achieves comparable video quality and semantic consistency with higher generation speed.

\begin{figure}[!t]
    \centering
    \includegraphics[width=0.42\linewidth]{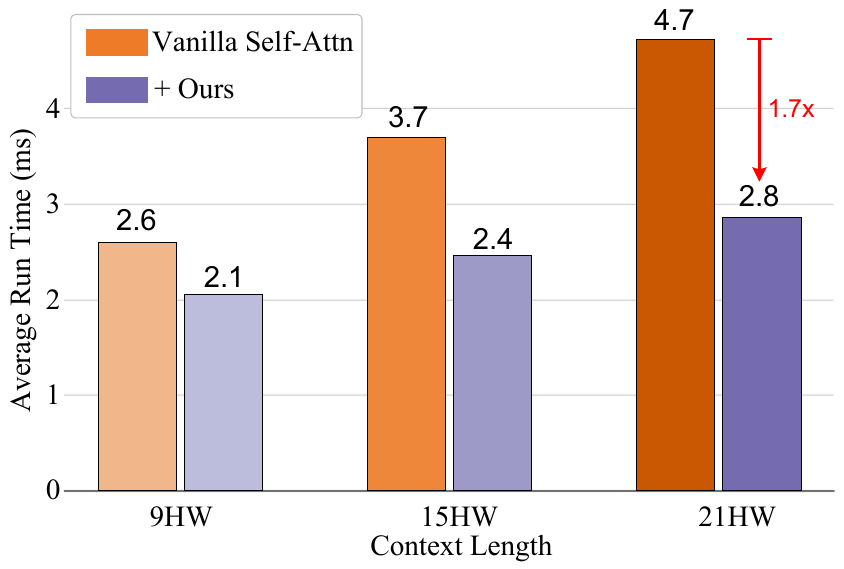}
    \vspace{-1mm}
    \caption{Runtime profiling on a single self-attention layer under different context lengths. $HW$ denotes the total number of visual tokens of one latent frame. The attention runtime is averaged across all layers of the model.}
    \label{fig:suppl_attention_time}
\end{figure}

\noindent \textbf{Speedup on Single Attention Layer.}
The main paper mainly demonstrates the speedup effect from end-to-end profiling. Notably, the proposed method accelerates only self-attention computations while sharing the same computational costs on other modules as the baseline model. To provide module-level acceleration results, ~\cref{fig:suppl_attention_time} shows the runtime of a single self-attention layer under different context lengths. Compared to previous baselines, which exhibit quadratic computational complexity with respect to visual token length, the acceleration effect of our \NAME becomes more pronounced as the sequence length increases. For instance, when self-attention processes interactions across 15 frames, the proposed \NAME achieves a 1.7$\times$ speedup.

\noindent \textbf{Results on 14B Models.}
In this section, we explore the effectiveness of the proposed method on large-scale autoregressive video diffusion. Specifically, we evaluate the proposed method on the RealTime-14B~\cite{krea_realtime_14b} model. RealTime-14B employs the Self Forcing training pipeline for post-training on the WAN2.1-14B~\cite{wan2025wan} model to obtain autoregressive video diffusion. The model contains a total of 1600 heads, and we set 800 heads as dummy heads. We evaluate the proposed method on 5-second short video generation on VBench~\cite{huang2024vbench} and 30-second long video generation on VBench-long~\cite{huang2024vbench++}. Results are presented in~\cref{tab:suppl_realtime14B}. It can be observed that the proposed method scales well to larger models. For instance, on VBench-long, pruning all KV cache of 50\% heads achieves even better total scores while delivering acceleration. The above experiments demonstrate the generalizability of the proposed method on large-scale models.

\begin{table}[!tb]
\centering
\caption{Quantitative comparison on cache compression ratios. We use Self Forcing as the baseline model and evaluate the KV cache reduction capabilities of different methods.}
\label{tab:suppl_compress_ratio}
\setlength{\tabcolsep}{3pt}
\scalebox{0.9}{
\begin{tabular}{@{}lcccccc@{}}
\toprule
methods     & cache\_reduction\_ratio & FPS   & speedup     & Semantic Score & Quality Score & Total Score \\ \midrule
Self Forcing & 100.0\%                  & 17.56 & 1.0$\times$ & 84.73 & 81.03&  84.00      \\
+ Infinipot-V & 16.7\%                  & 19.54 & 1.1$\times$ & 82.72          & 79.05         & 81.99       \\
+ R-KV        & 16.7\%                  & 18.41 & 1.1$\times$ & 83.46          & 79.60         & 82.69       \\
+ \textbf{\NAME}        & 27.8\%                  & 24.30 & 1.4$\times$ & 84.63          & 80.98         & 83.90       \\ \bottomrule
\end{tabular}%
}
\end{table}

\noindent \textbf{Comparison on Cache Compression Ratio.}
Since our approach works by compressing the length of the KV cache, it is interesting to explore how many caches are compressed. In~\cref{tab:suppl_compress_ratio}, we present the KV cache compression ratios of different methods. For the previous KV cache compression methods Infinipot-V~\cite{kim2025infinipot} and R-KV~\cite{cai2025rkv}, we set the cache budget to 1.5 frames, including 1 sink frame and 0.5 neighbor frames, which reduces the cache length to 16.7\% of the original baseline. For the proposed method, we consider the dummy \& non-dummy binary classification as a compression lower bound. This includes 1 packing frames for 50\% heads and 4 cached frames (1 sink + 3 neighbor frame) for the remaining 50\% heads, reducing the cache length to 27.8\% of the baseline. Notably, previous KV cache compression methods require recalculating token importance scores to select tokens at each AR step. This additional computational overhead undermines the benefits of shorter context length, ultimately resulting in suboptimal end-to-end speed. Furthermore, previous methods are not specifically designed for autoregressive video diffusion models, leading to noticeable performance degradation when affecting context-critical heads.  In contrast, the proposed \NAME requires only a single head classification per shot, followed by cache reduction based on predefined rules, significantly reducing additional overhead of the pruning algorithm. This enables our method to achieve faster runtime while preserving more tokens, striking a good balance between generation quality and speed.

\begin{figure}[!t]
    \centering
    \includegraphics[width=0.99\linewidth]{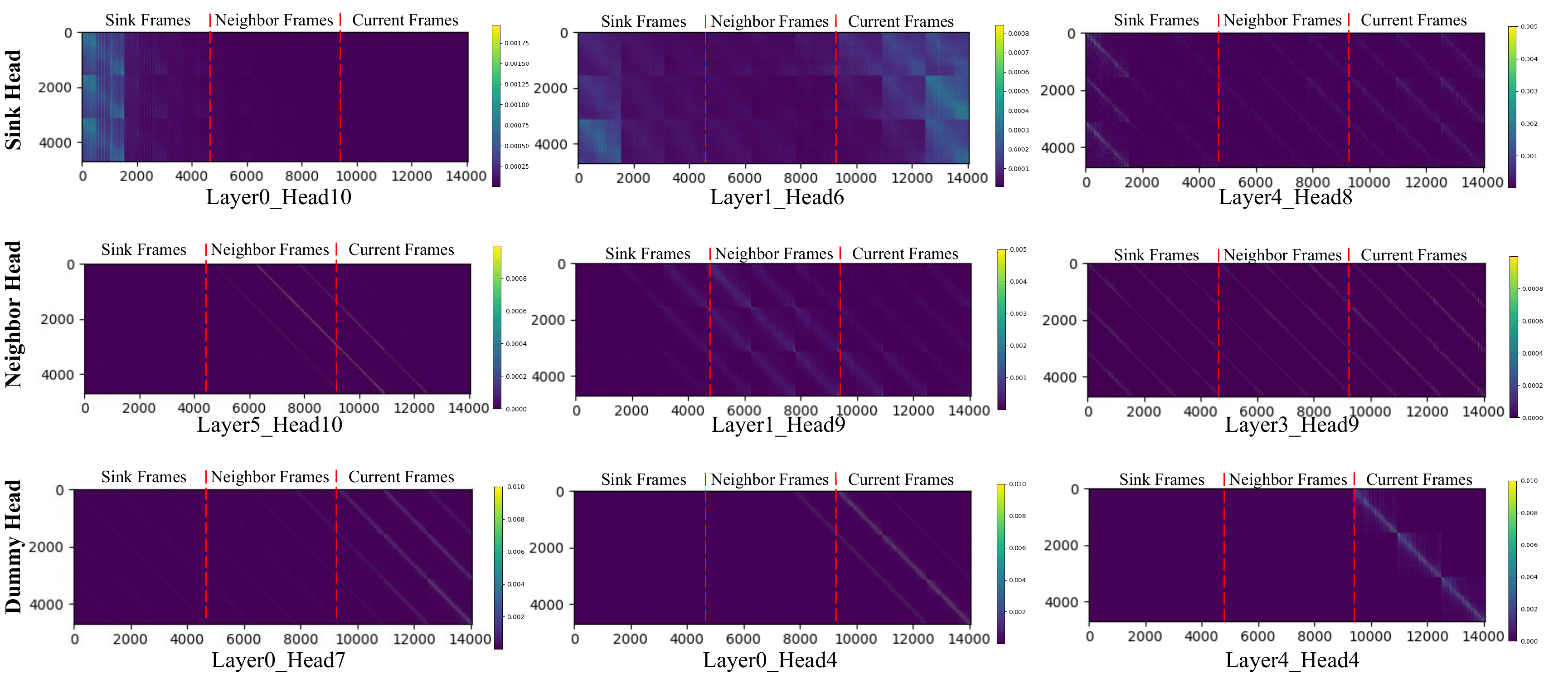}
    \vspace{-1mm}
    \caption{Attention maps visualization of sink, neighbor, and dummy heads. Zoom in for better effects.}
    \label{fig:suppl_attention_map}
\end{figure}

\noindent \textbf{Attention Map Visualization.}
In this section, we present the visualizations of attention maps. We employ the Self Forcing model as the base model, which utilizes the chunk-by-chunk generation scheme where each chunk comprises 3 latent frames. We select the third AR step because the equal lengths of the sink, neighbor, and dummy frames facilitate visualization. ~\cref{fig:suppl_attention_map} shows representative attention maps for the three head types: sink, neighbor, and dummy heads. As observed, different head types exhibit corresponding attention distribution patterns. For instance, the sink head shows significantly higher attention score to the sink frame compared to the other two head types, while the dummy head allocates most attention scores to the current frame. These visualizations further validate the proposed head-specific KV cache management strategy.

\begin{table}[!tb]
\centering
\caption{Quantitative comparison with RealTime14B model in 5-second short video generation on Vbench and 30-second long video generation on VBench-long.}
\label{tab:suppl_realtime14B}
\setlength{\tabcolsep}{3pt}
\scalebox{0.9}{
\begin{tabular}{@{}lccccc@{}}
\toprule
setups & duration & FPS & Quality & Semantic & Total Score \\ \midrule
RealTime14B~\cite{krea_realtime_14b} & 5s & 2.8 & 85.24 & 81.69 & 84.53 \\
+ \NAME (Ours) & 5s & 3.2 & 84.61 & 81.73 & 84.04 \\ \hdashline[0.5pt/2pt]
RealTime14B~\cite{krea_realtime_14b} & 30s & 2.8 & 84.23 & 72.91 & 81.97 \\
+ \NAME (Ours) & 30s & 3.2 & 84.56 & 76.42 & 82.94 \\ \bottomrule
\end{tabular}%
}
\end{table}

\section{Details of Long-context Generation}
\label{sec:suppl_long_context_video_gen}
In this section, we provide details on the experimental setup for long-context video generation.

\noindent
\textbf{Setups.}
In multi-shot video generation scenarios, it is easy for humans to evaluate the consistency of characters across multiple shots. However, designing automated quantitative evaluation protocols remains challenging. To this end, we design the following ``A-B-A'' like long-context video generation task. Specifically, we force the video generation model to generate 15s interactive videos driven by 3 text prompts $\mathcal{P}=\{P_1, P_2, P_3\}$, with each prompt controlling a 5s video segment. We use all prompts from VBench as $P_1$. To induce a shot transition in $P_2$, we prompt the Qwen2.5-72B-Instruct~\cite{team2024qwen2} model to generate a new shot that is different from $P_1$ to yield $P_2$. The related prompt is given in~\cref{fig:suppl_long_context_prompt}. Next, we set $P_3 = P_1$ to reconstruct the exact same scene. By evaluating the similarity between the first and the third video segments, we can automatically obtain quantitative scores that reflect the context utilization capability of different models.

\noindent \textbf{Metrics Calculation.}
Following the VBench~\cite{huang2024vbench} setup, we include the following video generation evaluation dimensions: subject\_consistency, background\_consistency, image\_quality, aesthetic\_quality, and overall\_consistency. The subject\_consistency evaluates consistency between frames by calculating similarity in the DINO~\cite{caron2021dino} feature space. Since the second video segment is prompted to generate a totally different transition scene, we extract the first and third segments to form a 10-second video clip for DINO similarity computation. The background\_consistency calculates similarity between two frames in the CLIP~\cite{radford2021CLIP} image encoder space. Similarly, we extract the first and third segments and compute similarity on the concatenated 10-second video clip. The image\_quality and aesthetic\_quality scores are obtained using the MUSIQ~\cite{beaumont2022aestheticpredictor} image quality predictor and LAION aesthetic predictor, respectively, and evaluated across all three video segments. Overall consistency is computed using video-text consistency calculated by ViCLIP~\cite{wang2023viCLIP} to reflect both semantic and stylistic consistency, where the text prompts contain distinct semantics and styles. Finally, we average the scores across all dimensions to obtain the total score.

\section{Limitation and Future Works}

Our \NAME effectively addresses the limited contextual utilization in current autoregressive video diffusion models by pruning redundant context from the dummy head. Nonetheless, our work can be further improved in the future in the following aspects.
First, the proposed method is currently a training-free pipeline. We note that post-training could potentially achieve further performance gains or improved compression rates. Specifically, we can first identify the dummy heads in pre-trained models using our proposed \NAME. Subsequently, by fine-tuning on a small dataset, we can completely remove the KV cache for dummy heads during training, forcing the model to concentrate its context aggregation capabilities on the few non-dummy heads. Due to computational constraints and the heavy workload involved, we reserve dummy head fine-tuning for future work.
Second, although we thoroughly analyze dummy heads in this paper, including their distribution across layers and commonality in existing models, further investigation into why they emerge in autoregressive video diffusion models remains an intriguing topic. This may involve analyzing training dynamics. Given that autoregressive video diffusion models represent a recent emerging generative paradigm, we believe future work will provide answers regarding the causes of dummy heads.
Third, since this work extensively explores the applicability of the proposed method to autoregressive video generation models, such as Diffusion Forcing, Self Forcing, and Rolling Forcing, further investigation into inference acceleration for other model categories, such as world models, represents a meaningful direction, and we leave it as future work.

\section{More Generation Results}
\label{sec:suppl_more_visual_results}
\begin{figure*}[!t]
    \centering
    \includegraphics[width=\linewidth]{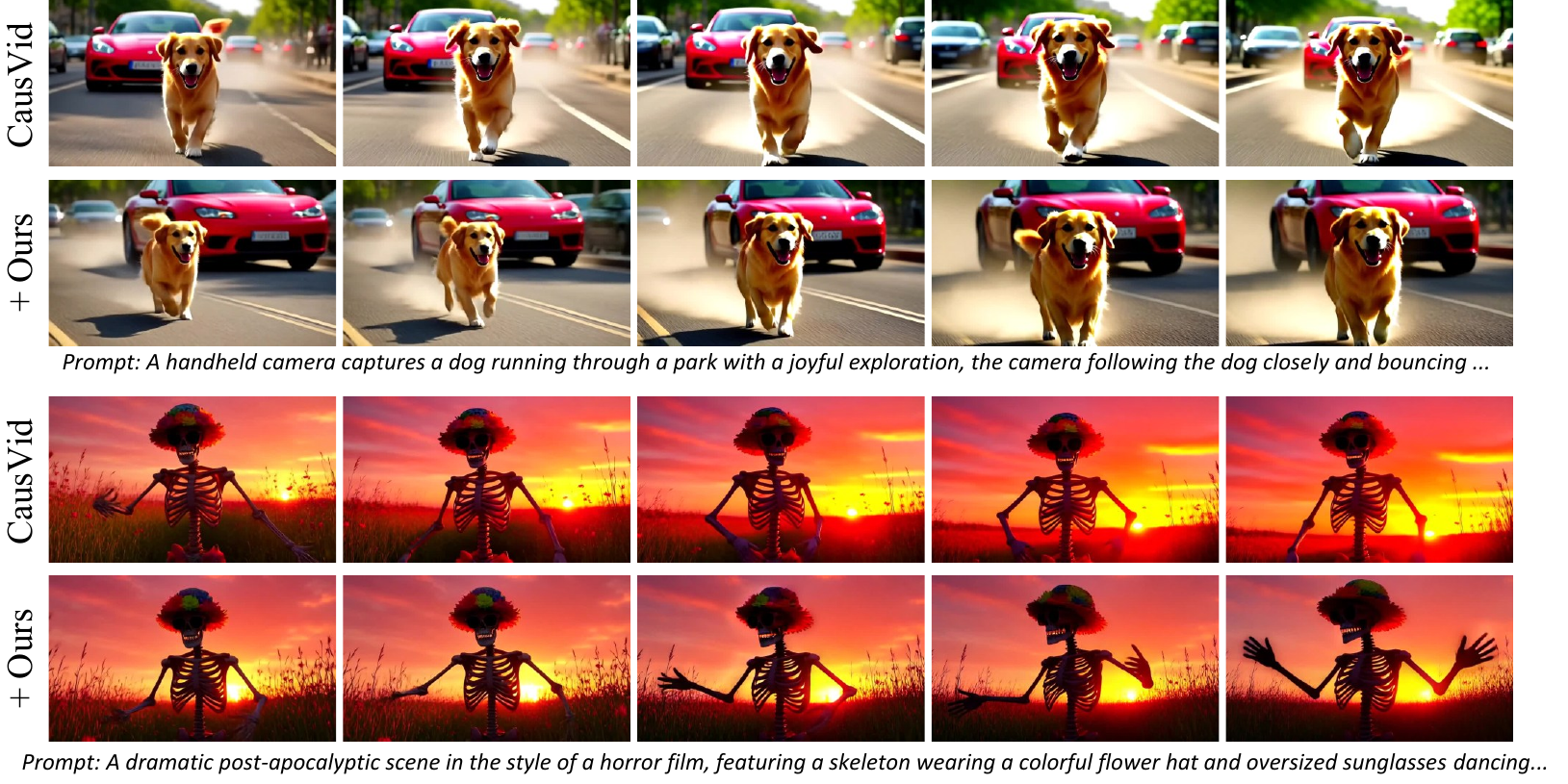}
    \vspace{-3mm}
    \caption{Quantitative comparison  between CausVid~\cite{yin2025causvid} and our \NAME. We prune 50\% heads' KV cache to serve as dummy heads in the proposed method.}
    \label{fig:suppl_CausVid_visual}
\end{figure*}

\begin{figure*}[!t]
    \centering
    \includegraphics[width=\linewidth]{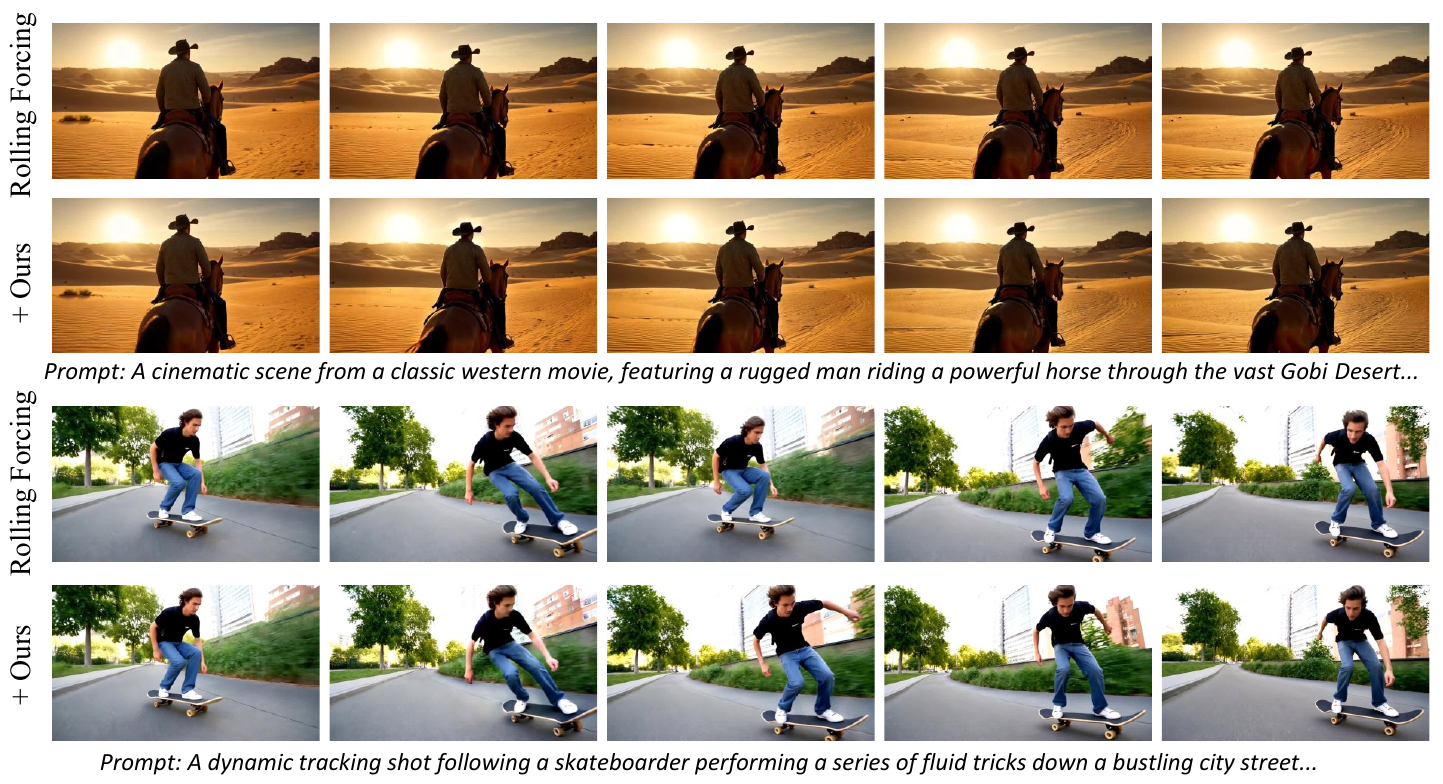}
    \vspace{-3mm}
    \caption{Quantitative comparison between Rolling Forcing~\cite{liu2025rollingforcing} and our \NAME. We prune 50\% heads' KV cache to serve as dummy heads in the proposed method.}
    \label{fig:suppl_rollingForcing_visual}
\end{figure*}

\begin{figure*}[p]
\centering
 \includegraphics[width=0.97\linewidth]{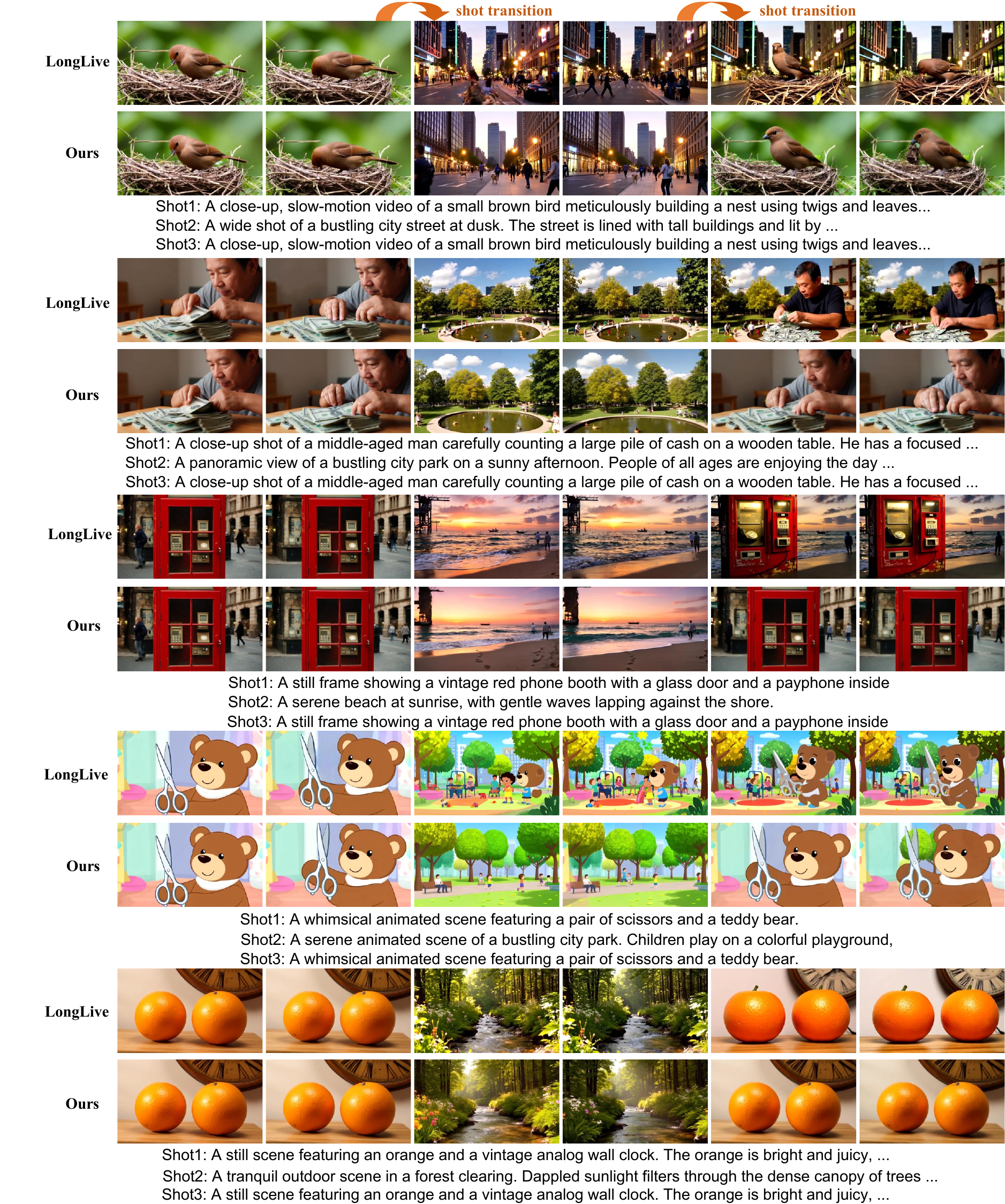}
 \caption{More quantitative comparison results in \textbf{long-context video generation} using context probing (Part1).}
 \label{fig:supp-context_probing_1}
\end{figure*}

\begin{figure*}[p]
\centering
 \includegraphics[width=0.97\linewidth]{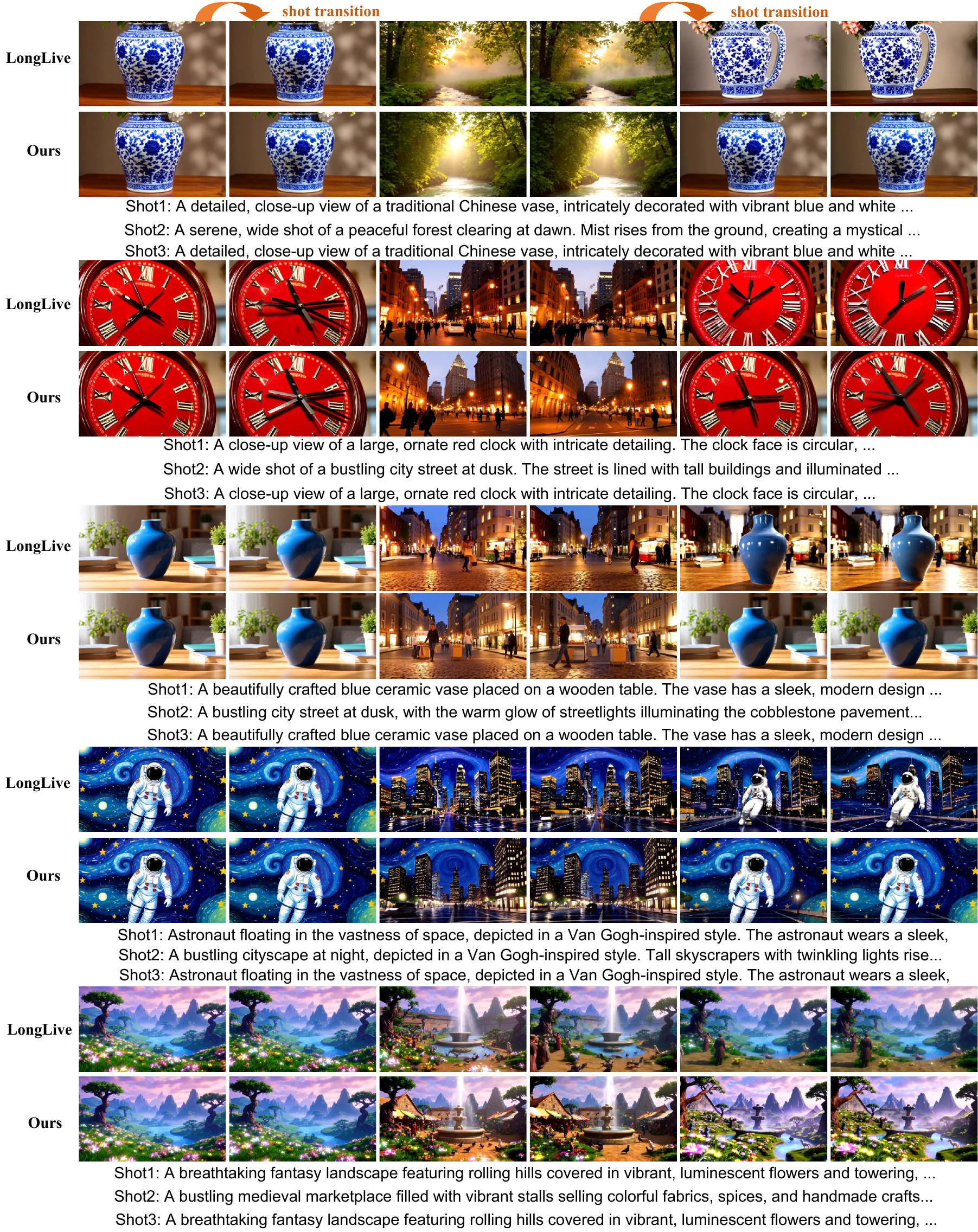}
 \caption{More quantitative comparison results in \textbf{long-context video generation} using context probing (Part2).}
 \label{fig:supp-context_probing_2}
\end{figure*}

\begin{figure*}[p]
\centering
\includegraphics[width=0.95\linewidth]{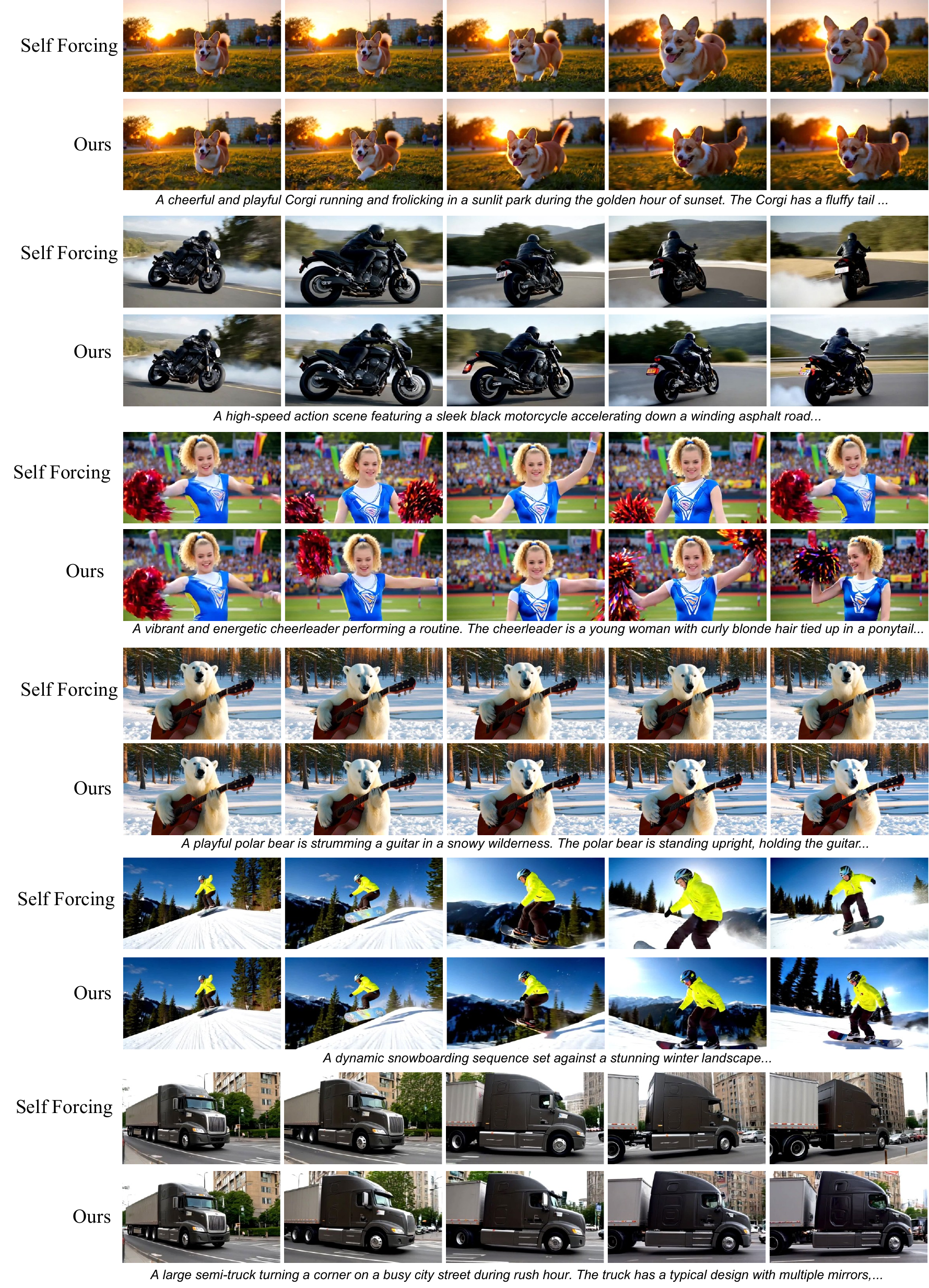}
\caption{More quantitative comparison results in \textbf{5s short video generation} task (Part1).}
\label{fig:supp-single5s_self_forcing1}
\end{figure*}

\begin{figure*}[p]
\centering
\includegraphics[width=0.95\linewidth]{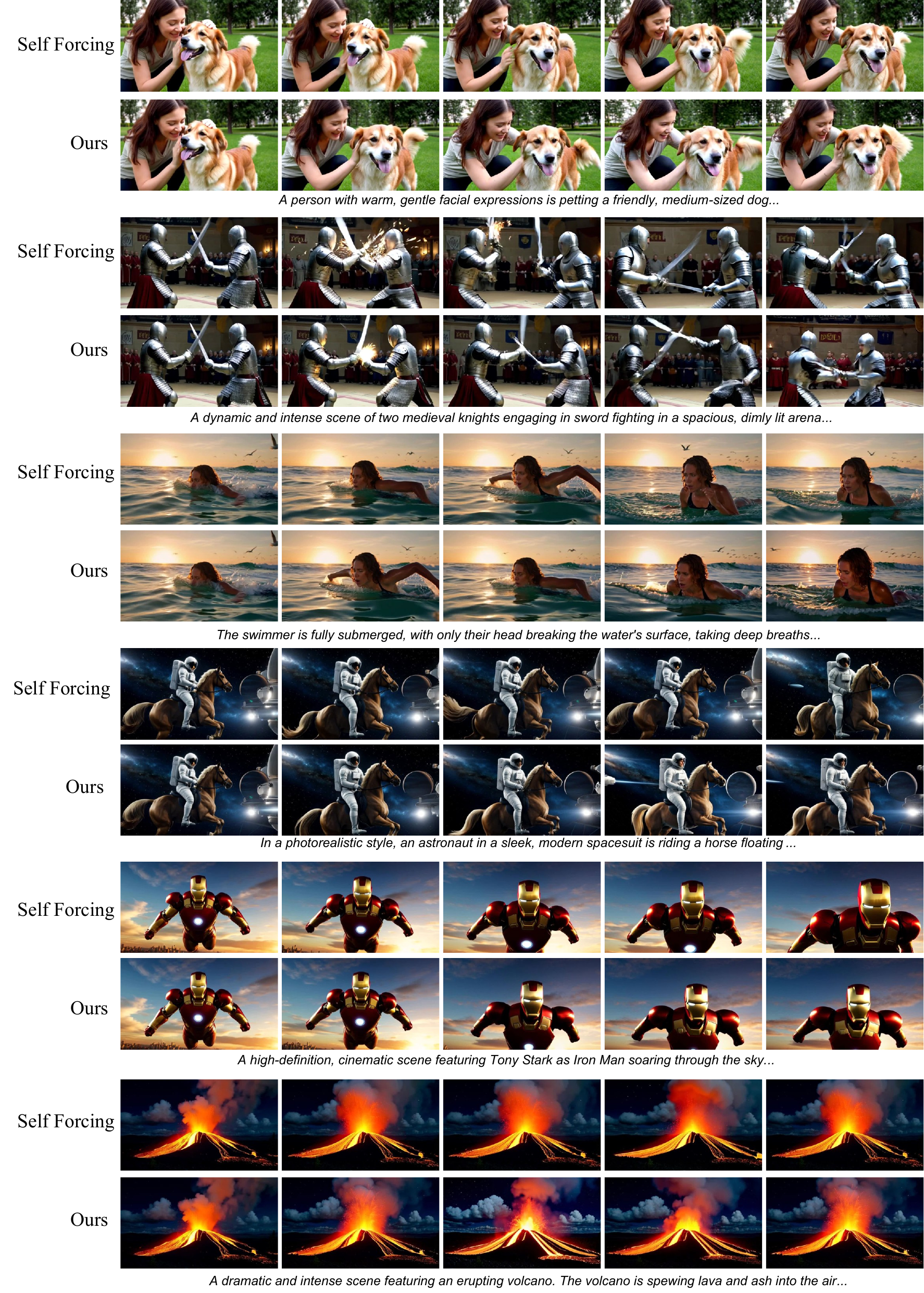}
\caption{More quantitative comparison results in \textbf{5s short video generation} task (Part2).}
\label{fig:supp-single5s_self_forcing2}
\end{figure*}

\begin{figure*}[p]
\centering
\includegraphics[width=0.96\linewidth]{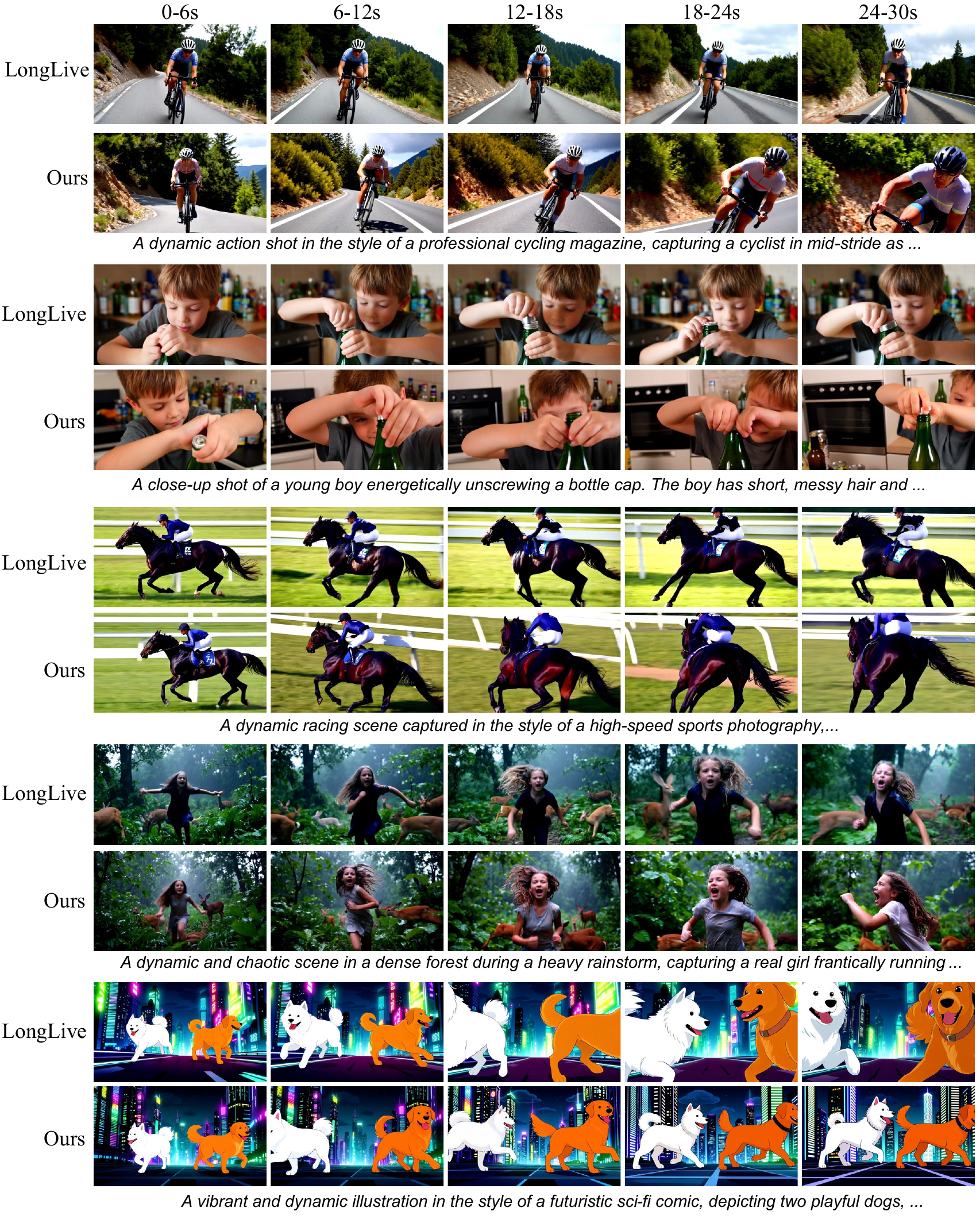}
\caption{More quantitative comparison results in \textbf{30s long video generation} task (Part1).}
\label{fig:supp-single30s_longlive1}
\end{figure*}

\begin{figure*}[p]
\centering
\includegraphics[width=0.96\linewidth]{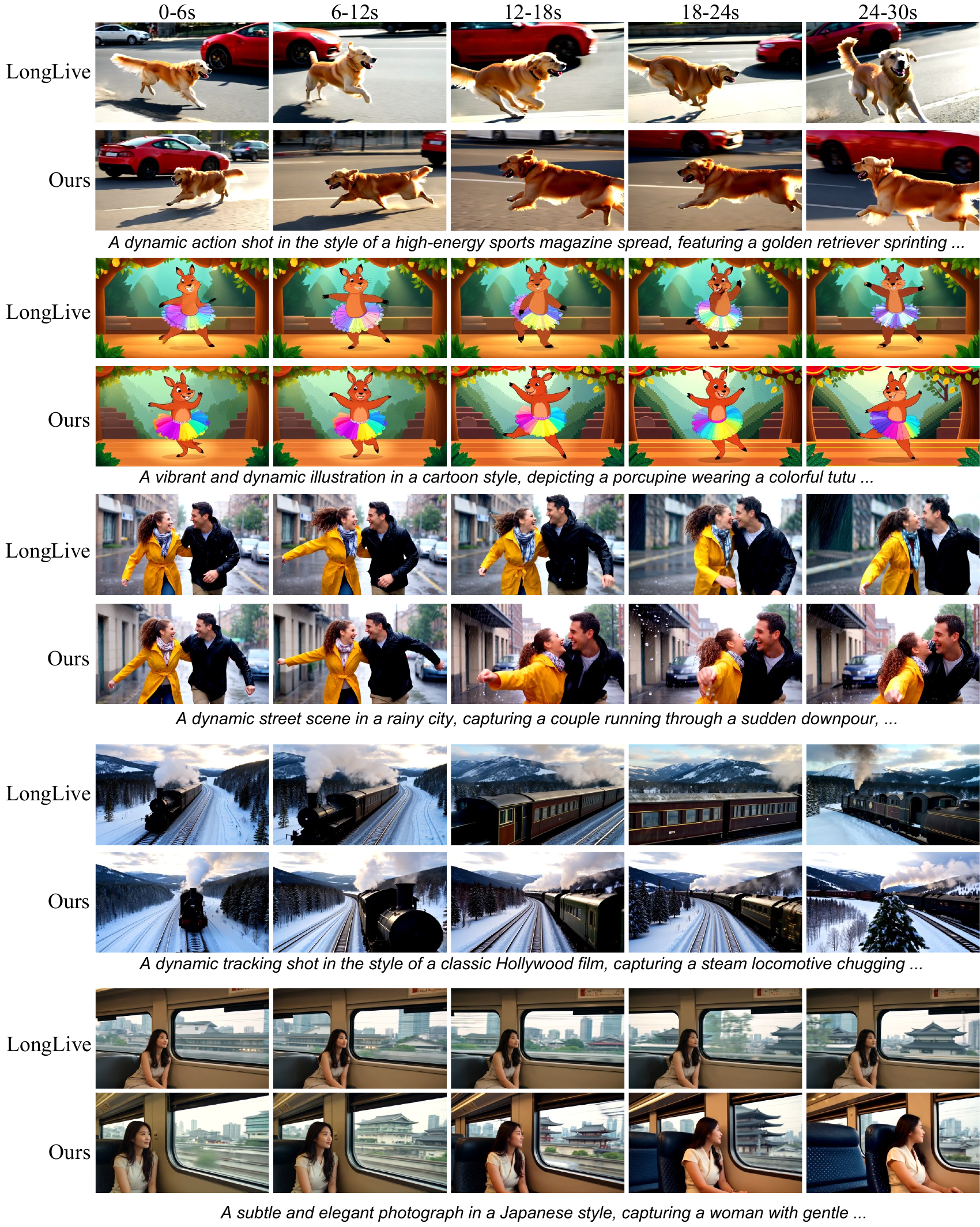}
\caption{More quantitative comparison results in \textbf{30s long video generation} task (Part2).}
\label{fig:supp-single30s_longlive2}
\end{figure*}

In this section, we provide more visual results, which are organized as follows:

\noindent \begin{itemize}
     \item In~\cref{fig:suppl_long_context_prompt}, we give the user prompt for generating the shot transition in our long-context video generation evaluation.
     \item In \cref{fig:suppl_CausVid_visual} and \cref{fig:suppl_rollingForcing_visual}, we give the results of applying \NAME to other autoregressive diffusion pipleines including Diffusion Forcing and Rolling Forcing, respectively.
     \item In \cref{fig:supp-context_probing_1} and \cref{fig:supp-context_probing_2}, we give more results on long-context video generation using the proposed context probing setup.
     \item \cref{fig:supp-single5s_self_forcing1} and \cref{fig:supp-single5s_self_forcing2} gives more quantitative comparison results on single prompt 5s short video generation.
    \item In \cref{fig:supp-single30s_longlive1} and \cref{fig:supp-single30s_longlive2}, we give more visualization on single prompt 30s long video generation task.
\end{itemize}

\clearpage
\begin{figure}[t]
\centering
\begin{Verbatim}[fontsize=\small, breaklines=true]
You are a professional video-prompt generation specialist. Your task is to
generate three text prompts for a multi-stage video generation system. These
prompts will be fed into a video generation model in sequence to evaluate its
scene transition behavior and its ability to preserve long-term visual memory.

Your input: you will receive one ORIGINAL_PROMPT describing the first shot.

Your output: return one python list containing three prompts in the format of
list(str), where

Prompt 1: exactly the ORIGINAL_PROMPT with no changes.

Prompt 2: a new prompt that introduces a required scene transition.

Prompt 3: a prompt that is exactly identical to the ORIGINAL_PROMPT.

Rules for Prompt 2:

- The setting or environment must be completely different from the
  ORIGINAL_PROMPT.

- Keep the mood, realism level, descriptive tone, and overall style consistent
  with the ORIGINAL_PROMPT.

- Do not use phrases that reference previous content, such as "still,"
  "as before," "continues," or similar.

- Use clear mid-level English and avoid rare or overly literary vocabulary.

Other rules:

- Prompt 2 must have 80-100 words.

- All prompts must be in English language.

- Prompt 1 and Prompt 3 must be exact copies of the ORIGINAL_PROMPT.

- Output format: ["PROMPT_1", "PROMPT_2", "PROMPT_3"]

Do not include any explanations, headings, markdown, or extra text.
Directly output the list of prompts.
\end{Verbatim}
\caption{System prompt used for shot transition in long-context video generation evaluation.}
\label{fig:suppl_long_context_prompt}
\end{figure}

\clearpage

\clearpage

\clearpage

\clearpage

\clearpage

\clearpage

\clearpage

\end{document}